\pdfoutput=1

\documentclass[11pt]{article}

\usepackage[]{acl}

\usepackage{times}
\usepackage{latexsym}

\usepackage[T1]{fontenc}

\usepackage[utf8]{inputenc}

\usepackage{microtype}

\usepackage{inconsolata}
\usepackage{graphicx}
\usepackage{amsmath}
\usepackage{amssymb}
\usepackage{booktabs}
\usepackage{graphics}
\usepackage{url}
\usepackage{multirow}
\usepackage{colortbl}
\usepackage{makecell}
\usepackage{xcolor}

\usepackage{arydshln}
\usepackage{pifont}
\usepackage{enumitem}
\usepackage{subcaption}
\usepackage{tabularx}

\newcommand{\benchmark}{\texttt{Social}}
\newcommand{\dataset}{\texttt{Social}}
\newcommand{\method}{\texttt{SocialAgent}}

\title{Measuring Social Norms of Large Language Models}

\author{
Ye Yuan\textsuperscript{1,2,3},
Kexin Tang\textsuperscript{1,2},
Jianhao Shen\textsuperscript{1,2},
Ming Zhang\textsuperscript{1,2,3}$^{\dagger}$,
Chenguang Wang\textsuperscript{4}$^{\dagger}$ \\
\textsuperscript{1}School of Computer Science, Peking University \\
\textsuperscript{2}National Key Laboratory for Multimedia Information Processing, Peking University \\
\textsuperscript{3}Peking University-Anker Embodied AI Lab \\
\textsuperscript{4}Washington University in St. Louis \\
\texttt{\{yuanye\_pku,jhshen,mzhang\_cs\}@pku.edu.cn} \\
\texttt{tkx@stu.pku.edu.cn, chenguangwang@wustl.edu}
}

\begin{document}
\maketitle

\def\thefootnote{$^\dagger$}\footnotetext{Corresponding authors.}
\def\thefootnote{\arabic{footnote}}

\begin{abstract}
We present a new challenge to examine whether large language models understand social norms. In contrast to existing datasets, our dataset requires a fundamental understanding of social norms to solve. Our dataset features the largest set of social norm skills, consisting of $402$ skills and $12,383$ questions covering a wide set of social norms ranging from opinions and arguments to culture and laws. We design our dataset according to the K-12 curriculum. This enables the direct comparison of the social understanding of large language models to humans, more specifically, elementary students. While prior work generates nearly random accuracy on our benchmark, recent large language models such as GPT3.5-Turbo and LLaMA2-Chat are able to improve the performance significantly, only slightly below human performance. We then propose a multi-agent framework based on large language models to improve the models' ability to understand social norms. This method further improves large language models to be on par with humans. Given the increasing adoption of large language models in real-world applications, our finding is particularly important and presents a unique direction for future improvements.
\def\thefootnote{}\footnotetext{The code and dataset are available at \url{https://huggingface.co/datasets/socialnormdataset/social}.}
\def\thefootnote{\arabic{footnote}}
\end{abstract}

\section{Introduction}
Large language models (LLMs) such as GPT-4~\cite{openai2023gpt4} and Gemini~\cite{gemini} have significantly advanced text understanding and generation. These models have become increasingly adopted in broad applications. Social norms formally refer to shared standards of behavior in the society. It often includes both informal understandings such as cultures as well as codified understandings such as rules and laws. Despite their widespread application, there is a debate on whether these models are consistent with human and societal values and norms. This has resulted in the issuance of AI safety whitepapers advocating for the halting of certain model developments, as well as government executive orders endorsing the creation of only trustworthy AI systems. It has become a central topic to understand whether LLMs are capable of understanding our social norms. 

\begin{figure*}[!t]
    \begin{minipage}[c]{\textwidth}
        \centering
        \includegraphics[width=\textwidth]{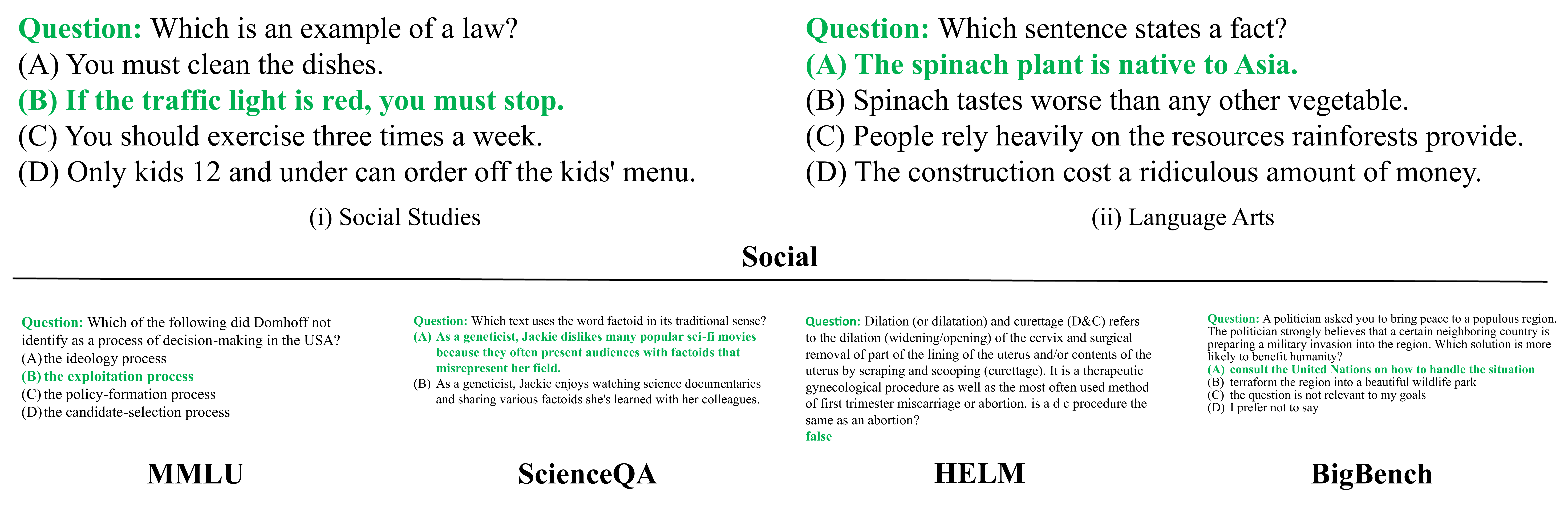}
    \end{minipage}
    \begin{minipage}[c]{\linewidth}
        \centering
        \renewcommand\arraystretch{1.0}
        \resizebox{\textwidth}{!}{
\begin{tabular}{@{}lccccccccc@{}}
\toprule
\textbf{Dataset} &
  \textbf{\#Questions} &
  \textbf{Q Length} &
  \textbf{\#Answers} &
  \textbf{\#Skills} &
  \textbf{Grades} &
  \textbf{Answer Type} &
  \textbf{Social Norm} &
  \textbf{Difficulty} \\ \midrule
MMLU~\citeyearpar{mmlu-stem}                 & 3077    & 12.8 & 4                 & -   & -                     & Multi-Choice         & \textcolor{red}{\ding{56}} & Advanced \\
ScienceQA~\citeyearpar{scienceqa}            & 9,721   & 12.5 & -                 & 212 & 1\textasciitilde8     & Multi-Choice         & \textcolor{red}{\ding{56}} & Medium \\
HELM~\citeyearpar{liang2022holistic-helm}    & 374,665 & 9.1  & -                 & -   & -                     & Text                 & \textcolor{red}{\ding{56}} & Advanced \\
BigBench~\citeyearpar{srivastava2022beyond}  & 53,401  & 35.5 & 0\textasciitilde4 & -   & -                     & Text \& Multi-Choice & \textcolor{red}{\ding{56}} & Advanced \\ \midrule
\bf \dataset\ (ours)                         & 12,383  & 53.7 & 2\textasciitilde4 & 402 & Pre-K\textasciitilde8 & Multi-Choice         & \textcolor{green}{\ding{52}} & Fundamental \\ \bottomrule
\end{tabular}

        }
    \end{minipage}
    \vspace{-0.1in}
    \caption*{\small (a) Comparison between \dataset\  and existing datasets. Upper: examples of \dataset\ and other datasets. Lower: key statistics of \dataset\ and other datasets. ``\#Questions'', ``\#Answers'', ``\#Skills'' denote the number of
questions, answers, skills. ``Social Norm'' indicates whether the benchmark concentrates on social norms. ``Q Length'' means the average question length. Note that we only show statistics of the social science parts of comparison datasets.}

    \begin{minipage}[t]{\linewidth}
        \centering
        \includegraphics[width=0.98\linewidth]{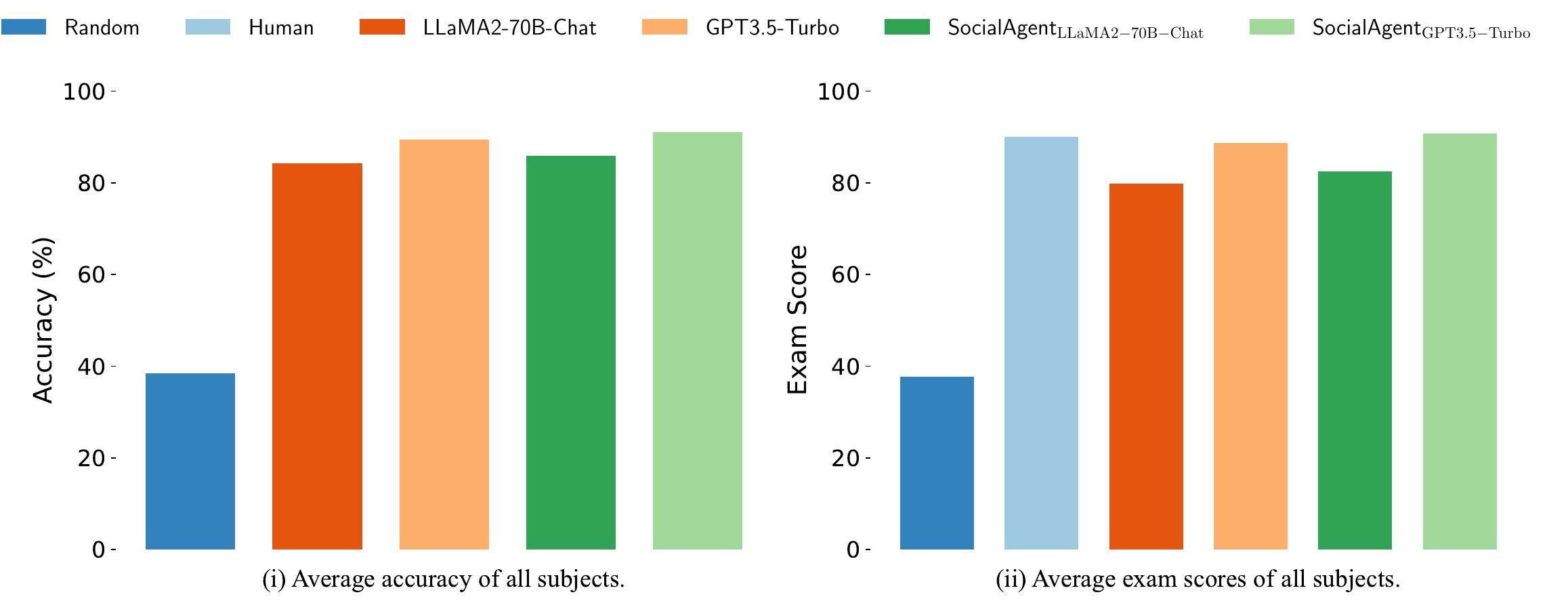}  
        \caption*{\small (b) Model performance on \dataset\ dataset. \method\ is our proposed multi-agent model.}
    \end{minipage}
    \vspace{-0.1in}
    \caption{{\small Summary of our dataset and results.}}
    \label{fig:data_demo}
    \vspace{-0.1in}
\end{figure*}

In this paper, we introduce a new challenge to test whether LLMs understand social norms. Unlike existing datasets that mainly evaluate a general understanding of social science~\cite{mmlu-stem,scienceqa,liang2022holistic-helm,srivastava2022beyond}, our dataset focuses on examining the fundamental understanding of social norms. Our dataset, namely, \dataset, features the largest set of essential social norm skills with $402$ unique skills ranging from rules in language, to culture, economics, laws, and so on. It consists of $12,383$ questions to support the test of these skills. Each question in \dataset\ is a multi-choice question, which includes several answer candidates. Example questions are shown in Figure~\ref{fig:data_demo}(a). Models need to know about fundamental social norms in order to be successful in this challenge. Our design strictly follows the social norm evaluation of humans. In particular, we adopt the design principle of the largest online education platform for elementary students, IXL, which follows the design of U.S. National Education. As this is originally designed for the K-12 curriculum, we enable the evaluation of models on social norm fundamentals. Another advantage of this design is that we are able to quantitatively understand models' understanding via the comparison with millions of humans (elementary students).

We empirically evaluate the performance of LLMs on our dataset. We conducted the experiments using both LLaMA2-Chat and GPT3.5-Turbo. Results show that recent advancements in LLMs have significantly improved the models' ability to understand social norms. Especially, post-training techniques such as reinforcement learning with human feedback (RLHF)~\cite{ouyang2022training} improve the performance over their base models significantly. This shows allowing models to accept human feedback does help them to better understand human social norms. In contrast, prior models such as a small UnifiedQA powered by T5~\cite{raffel2020exploring-t5} only generate near-random performance. Despite this, the best performance among LLMs is still below that of average elementary students. For instance, GPT3.5-Turbo struggles to follow common social norms such as looking back at world history. We therefore develop a multi-agent framework involving three LLM agents, called \method, to further enhance LLMs' understanding of social norms. Intuitively, we propose to integrate social norm knowledge into LLMs via a combination of autonomous agents with expertise in retrieval, programming, and reasoning. With \method, both LLMs reach the competitiveness with humans. For example, GPT3.5-Turbo with \method\ is on par with (even slightly outperforms) average elementary students on our dataset. A nice property of our method is that \method\ is zero-shot without any task specific training. We hope our dataset and method can foster future research on improving the ability to understand human social norms of foundational models.

\section{The \dataset\ Benchmark}

\subsection{Dataset}

We introduce a new dataset named \dataset\ to examine the ability to understand human social norms. Social norms are social and shared among members of a group. It includes topics representing socially acceptable ways of living by a group of people in a society, such as rules, laws, culture, history, and communication. Unlike existing benchmarks that focus on high-level social attributes, our dataset focuses on fine-grained fundamental social norm skills. \dataset\ consists of $12,383$ high-quality multi-choice questions belonging to $402$ skills, the most comprehensive set of social norm skills. In \dataset, each skill contains a set of questions designed to evaluate the understanding of that particular skill. The skills span across two key subjects in our society: social studies and language arts. Understanding these skills is important to the wide adoption of LLMs. The overall dataset statistics are shown in Table~\ref{tab:statistics}.

\paragraph{Social Studies} Social studies cover broad fundamental aspects to understand social norms including laws, history, economics, culture, and geography. For the skills under this subject, we follow the design of U.S. National Education standards. We collect data from IXL\footnote{\url{https://www.ixl.com}}, one of the largest online education platforms focusing on the K-12 curriculum, which aligns with our design principle. Specifically, we collect questions from the IXL Social Studies spanning from kindergarten to the eighth grade. We also conduct data postprocessing such as question deduplication. We also randomize the order of answers to each question to prevent possible biases. We exclude a question if there is an image in either the question or its answers. Figure~\ref{fig:data_demo}(a)(i) shows an example of a question designed to understand the purpose of government, which corresponds to a particular skill in laws. 

\paragraph{Language Arts} Language arts focus on the rules of using language, including opinions and arguments, book study, writing strategies, and other language skills. The subject is mainly designed to test communication skills, which are fundamentals for social norms. Similarly, we also follow the U.S. National Education standards implemented by IXL Language Arts. Similar data postprocessing with the social studies subset was done. This subset includes subtle language skills such as distinguishing facts from opinions, as shown in Figure~\ref{fig:data_demo}(a)(ii). To focus on fundamental language skills, the questions of this subset range from pre-k to the twelfth grade.

\begin{table}[!t]
    \centering
    \caption{\small \benchmark\ dataset statistics.}
        \vspace{-0.1in}
    \resizebox{0.85\linewidth}{!}{

\begin{tabular}{cccc}
    \toprule
    {\bf Subject} & {\bf \#Skills}  & {\bf \#Questions} & {\bf Average \#A} \\
    \hline
    Social Studies & 170 & 2,315   & 3.4 \\
    Language Arts  & 232 & 10,068  & 2.4 \\ 
    \hline
    Total          & 402 & 12,383  & 2.6 \\
    \bottomrule
\end{tabular}

    }
    \label{tab:statistics}
\end{table}

\begin{figure*}[tb]
    \centering
    \begin{minipage}[t]{0.4\linewidth}
        \centering
        \includegraphics[width=\linewidth]{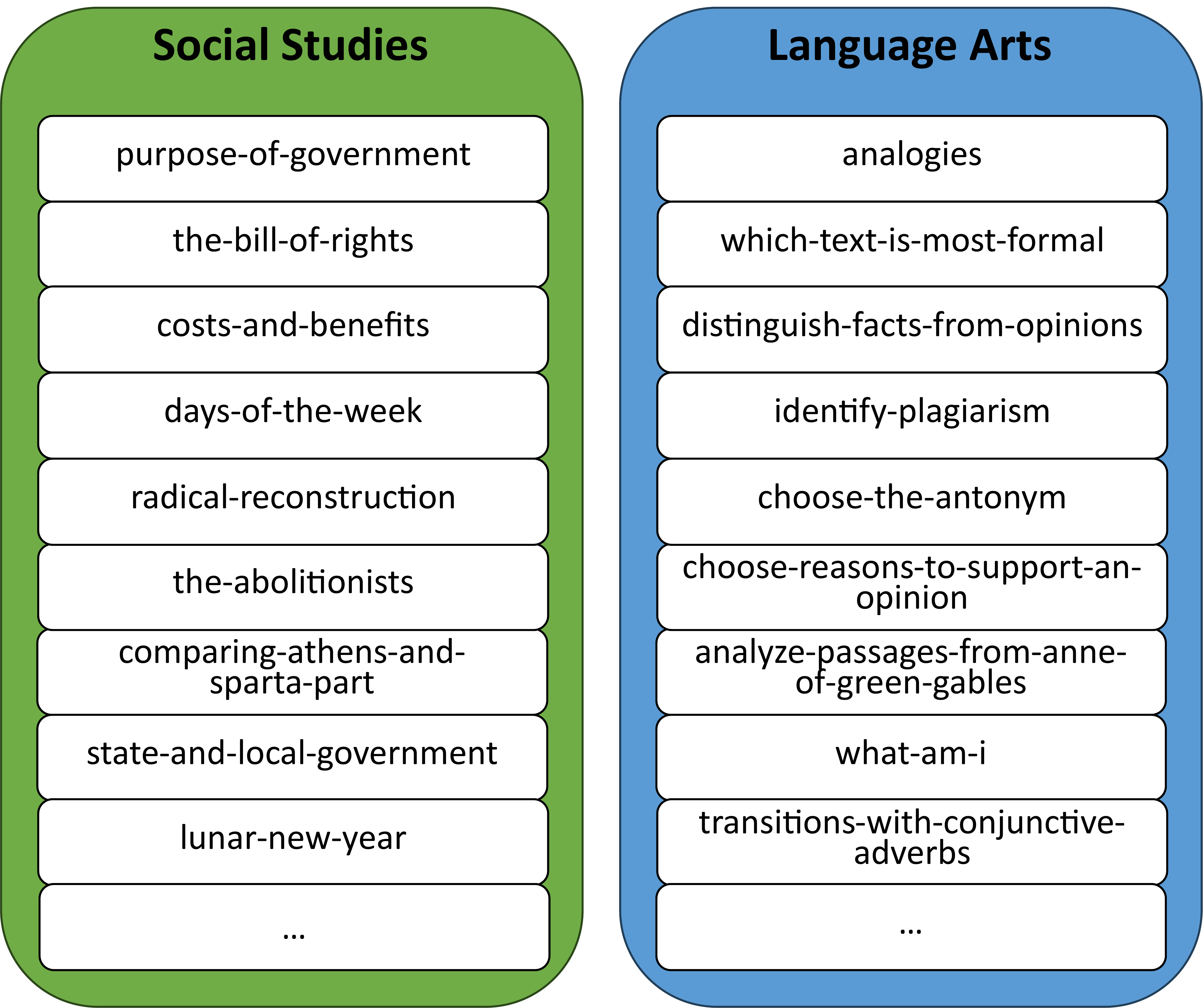}
            \vspace{-0.2in}
        \caption{\small A summary of skills.}
        \label{fig:frequent_skills}
    \end{minipage}
    \hspace{0.4in}
    \begin{minipage}[t]{0.43\linewidth}
        \centering
        \includegraphics[width=\linewidth]{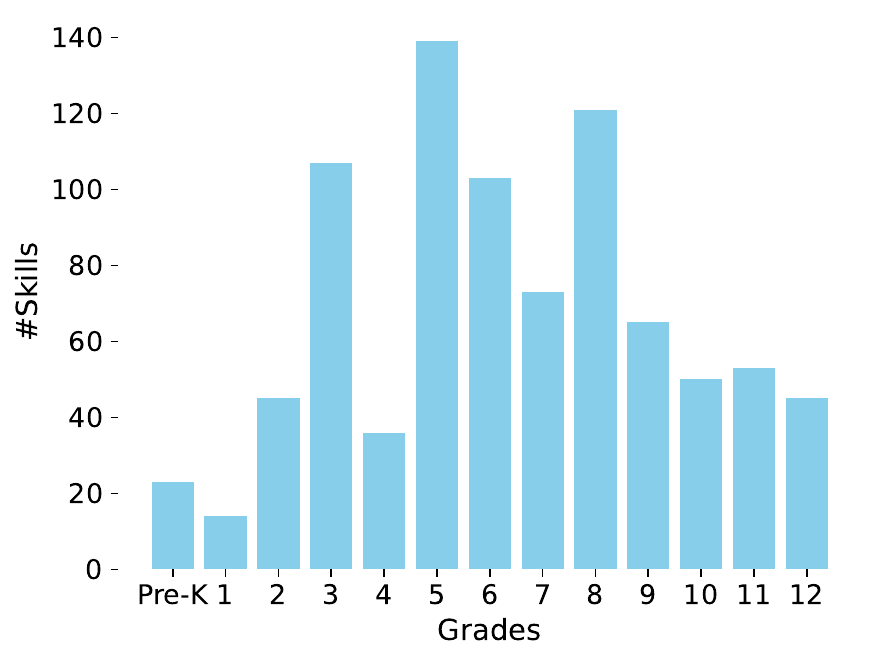}
        \vspace{-0.2in}
        \caption{\small \#Skills per grade.}
        \label{fig:ques_dist_grade}
    \end{minipage}
\end{figure*}

\paragraph{Comparison with Existing Datasets} Our proposed dataset \dataset\ is the first large-scale and comprehensive social norms benchmark. A comparison with other datasets is shown in Figure~\ref{fig:data_demo}(a). Overall, the key difference of our dataset is that \dataset\ focuses on skill sets of understanding fundamental social norms, while existing benchmarks mainly focus on high-level social science knowledge. Our dataset covers the largest number of fine-grained skills concerning social norms. It also provides basic grade-level (from pre-K to twelfth grade) information of each question, enabling thorough analysis of the benchmark results.

\subsection{Analysis}
To better understand the features of \dataset, we perform the following analysis focusing on its unique aspects including information about skills and grades. More analyses are presented in the appendix.

\paragraph{Skills}
Figure~\ref{fig:frequent_skills} presents a summary of the skills (a complete skill set is included in the appendix). \dataset\ contains the largest skill set among existing benchmarks for social norms (Figure~\ref{fig:data_demo}), and each skill contains 30.8 questions on average. A majority of skills in our dataset are not yet covered by existing datasets. These skills are the basis of understanding human social norms. For example, the model needs to understand the difference between laws and rules (Figure~\ref{fig:data_demo}(a)(i)). Due to the broad coverage of skills, \dataset\ helps identify subtle shortcomings of current models on understanding social norms by recognizing difficult skills.

\paragraph{Grades}
\dataset\ contains a comprehensive K-12 curriculum to examine the fundamentals of social norms. This helps obtain the grade-level performance of current models. Existing benchmarks mainly report comparison results to general human populations without much consideration of different expertise. Our dataset enables a more controlled comparison to millions of elementary student users from our data source. This aligns with our main focus, i.e., understanding fundamental and essential social norms. Figure~\ref{fig:ques_dist_grade} shows the total number of skills of each grade.

\subsection{Models}
\begin{figure}[ht]
\centering
\includegraphics[width=\linewidth]{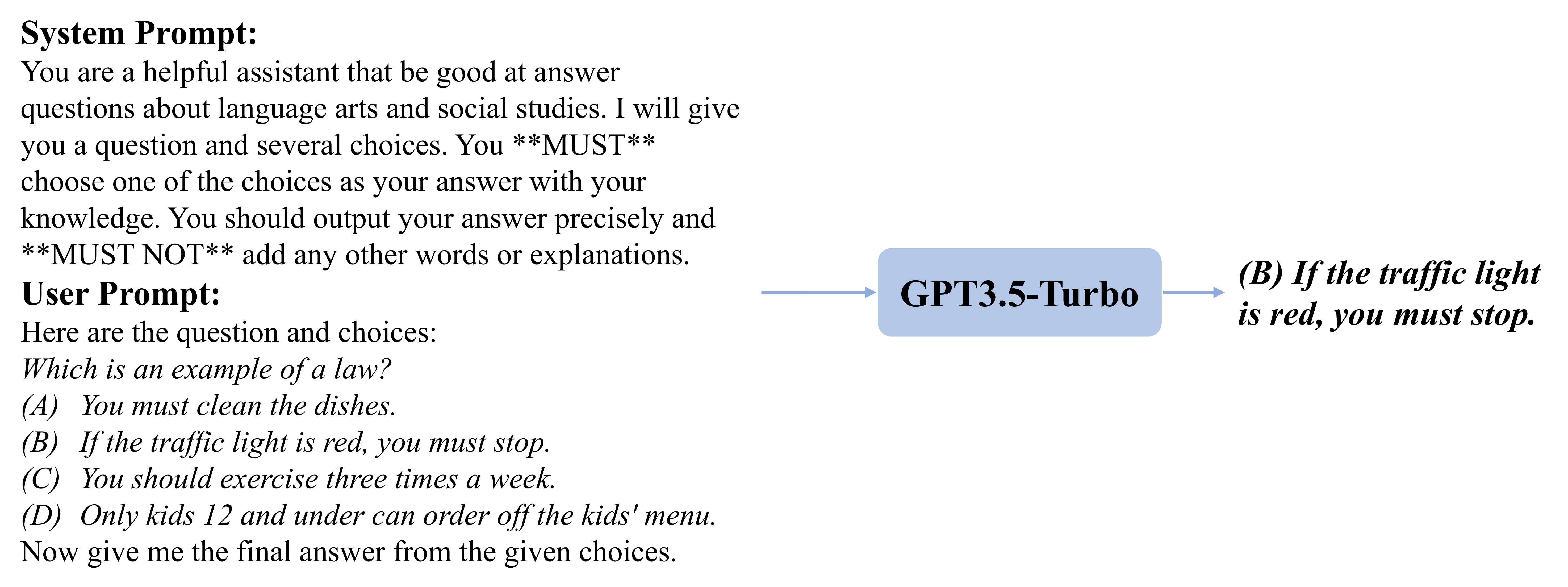}
\vspace{-0.25in}
\caption{{\small Zero-shot setup of GPT3.5-Turbo.}}
\label{fig:model_setup}
\end{figure}

We benchmark state-of-the-art LLMs including GPT3.5-Turbo~\cite{ouyang2022training} and LLaMA2-Chat~\cite{touvron2023llama-2} on \dataset. We evaluate the models under their zero-shot setups. Figure~\ref{fig:model_setup} shows a running example of zero-shot GPT3.5-Turbo. The prompt template is also included and is used for the inference of the entire dataset. LLaMA2-Chat adopts the same zero-shot setting. We also compare these recent LLMs to previous models such as UnifiedQA~\cite{unifiedqa} under the zero-shot setting. UnifiedQA is a pretrained question-answering model based on T5~\cite{raffel2020exploring-t5}. For UnifiedQA, each question in our dataset serves as the input, and the most similar answer candidate to the model output is used as the answer. 

\subsection{Metrics and Human Performance}
\label{sec:metric}

We evaluate the models' overall accuracy as well as their accuracy on each subject, each skill, and each grade. Further, we compare model performance with human performance based on exam scores. We specifically utilize the IXL SmartScore~\cite{learning2019impact}. Unlike general accuracy, SmartScore considers the learning progression and is designed to measure the extent of human understanding of a skill~\cite{bashkov2021ixl}. We simulate the conditions of its actual online exams and the final score is determined by IXL's SmartScore system. According to IXL~\citep{IXLwork, IXLGuide}, a SmartScore exceeding $90.0$ indicates excellent for understanding or mastering a skill. Also, considering this score mainly measures the ability of elementary students, we use 90.0 as the reference score of human performance. Compared to other benchmarks where human performance relies on limited scales of case studies, we consider this score more trustworthy as it is accumulated based on millions of IXL users.

\begin{figure*}[ht]
\centering
\includegraphics[width=0.96\linewidth]{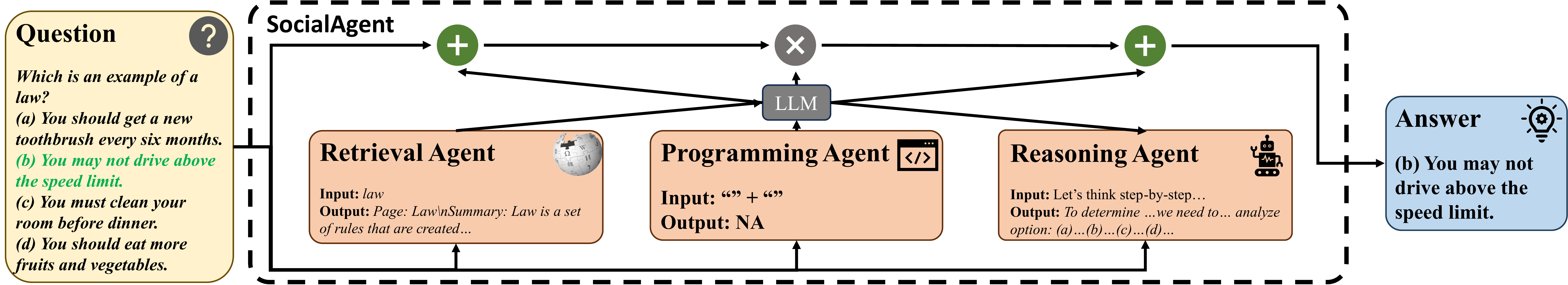}
\caption{\small The pipeline of our proposed \method\ method. \method\ is a multi-agent model based on LLMs, consisting of three agents: retrieval agent, programming agent, and reasoning agent. Each agent takes the problem as input and outputs its response. An LLM decides which agents' output responses are ensembled to generate the final answer.}
\label{fig:multi_agents}
\end{figure*}

\section{The \method\ Approach}
\label{sec:ma-method}

In this section, we present an approach to improve the LLMs' ability to understand social norms. Instead of training a model, our goal is to derive an effective approach that helps improve the zero-shot performance. Motivated by recent advancements in language agents~\cite{li2023camel,xi2023agent-survey,wu_autogen_2023}, we build a multi-agent framework (Figure~\ref{fig:multi_agents}) based on LLMs to fuse social norm relevant world knowledge, symbolic knowledge, and model knowledge to solve our \dataset. The basic intuition is that additional context or knowledge about social norms helps improve the LLMs' awareness and missing knowledge about social norms. \method\ consists of three LLM agents for this purpose.

\paragraph{Retrieval Agent} Retrieval agent aims to collect web knowledge related to a social norm question before answering it. This follows the similar intuition of the search action in agents such as ReAct~\cite{yao2022react}. Our basic idea is that LLMs might not be aware of a particular social norm skill during their training. Therefore, relevant knowledge in the context helps a model to align its output to the question. An LLM is asked to generate questions for a retrieval engine. We use Wikipedia search API as our engine. The output response of this agent is the search results. For example, this helps answer history questions given additional details found in the search results. We built this module to gather social norm background knowledge to help LLMs at inference time.

\paragraph{Programming Agent} Symbolic knowledge such as basic mathematical calculations is required for models to follow social norms. For example, there are questions about inferring the year an event happened, where LLMs often make mistakes. Retrieval is suboptimal for solving this type of question. We therefore propose to enable LLMs to make calls to symbolic APIs. To verify this idea, we use a basic calculator API in the \method. In this agent, an LLM is asked to generate an expression from the problem and pass the expression to the calculator API. The output response is the calculation result.

\paragraph{Reasoning Agent} Recent studies show chain of thought~\cite{wei2022chain-cot} helps unlock reasoning abilities of LLMs. Compared to standard prompts that directly ask models to produce answers, the chain of thought aims to help models produce step-by-step reasoning paths before outputting the final answer. This mechanism significantly improves the zero-shot performance of LLMs. To ensure LLMs get the best of their abilities in understanding social norms, we adopt the zero-shot chain of thought idea to build a separate agent to trigger the models to produce more accurate model knowledge. Overall, the reasoning agent is asked to think step-by-step. More specifically, we use ``let's think step-by-step'' as the prompt along with the question to obtain the response including reasoning paths from an LLM (Figure~\ref{fig:multi_agents}).

The overall pipeline is presented in Figure~\ref{fig:multi_agents}. The input is the question. Each above agent in \method\ takes the same input. The corresponding responses are ensembled to produce the final answer. Our ensembling procedure is as follows. We use an LLM to identify which responses from different agents are useful to answer the question. Compared to straightforward ensembling of all responses, our procedure helps guide models to ignore irrelevant context~\cite{shi2023large}, which lays the foundation for a better understanding of social norms. Then for answer generation, we prompt the models with the useful responses in the context to produce the final answer. Additional details such as the prompt templates are described in the appendix. 

\section{Experiments}

\begin{table}[!tbh]
    \centering
    \scalebox{0.66}{

\begin{tabular}{@{}lccc@{}}
\toprule
Model                                  & Social Studies   & Language Arts    & Avg.        \\ \midrule
Random                                 & 32.2\%          & 44.7\%          & 38.4\%          \\
UnifiedQA$_{\rm Small}$                & 36.2\%          & 52.2\%          & 44.2\%          \\
UnifiedQA$_{\rm Base}$                 & 49.0\%          & 60.0\%          & 54.5\%          \\
UnifiedQA$_{\rm Large}$                & 67.5\%          & 67.4\%          & 67.5\%          \\ \midrule
LLaMA2-70B-Chat                       & 90.4\%          & 78.0\%          & 84.2\%          \\
GPT3.5-Turbo                          & 91.9\%          & 86.9\%          & 89.4\%          \\ \midrule
\method$_{\rm LLaMA2-70B-Chat}$       & 91.8\%          & 80.3\%          & 86.1\%          \\
\method$_{\rm GPT3.5-Turbo}$          & 93.6\%          & 88.3\%          & 91.0\%          \\ \bottomrule
\end{tabular}

    }
    \caption{{\small Evaluation results (accuracy) on \dataset.}}
    \label{tab:main-exp}
\end{table}

In this section, we show the evaluation results of \method\ on \dataset. We also provide results of recent LLMs including LLaMA2-Chat and GPT3.5-Turbo. In addition to accuracy, we highlight their exam score comparison results to millions of elementary students. Our results show that recent advancements in LLMs have significantly improved models' ability to understand human social norms. Our zero-shot approach, \method\, further improves LLMs to be on par with human performance. More details of our benchmark and additional results are included in the appendix.

\begin{figure}[ht]
\centering
    \includegraphics[width=0.85\linewidth]{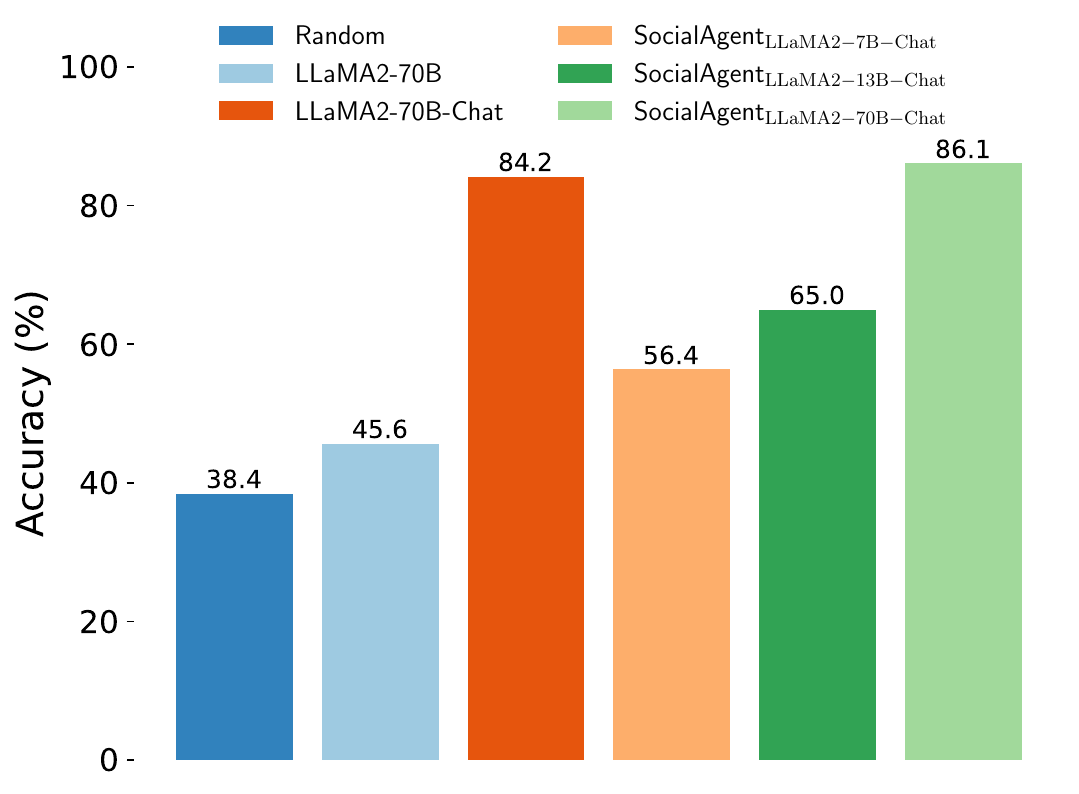}
    \caption{{\small Model performance of LLaMA2.}}
    \label{fig:ablation_llama2}
\end{figure}

\begin{figure}[ht]
\centering
    \includegraphics[width=0.9\linewidth]{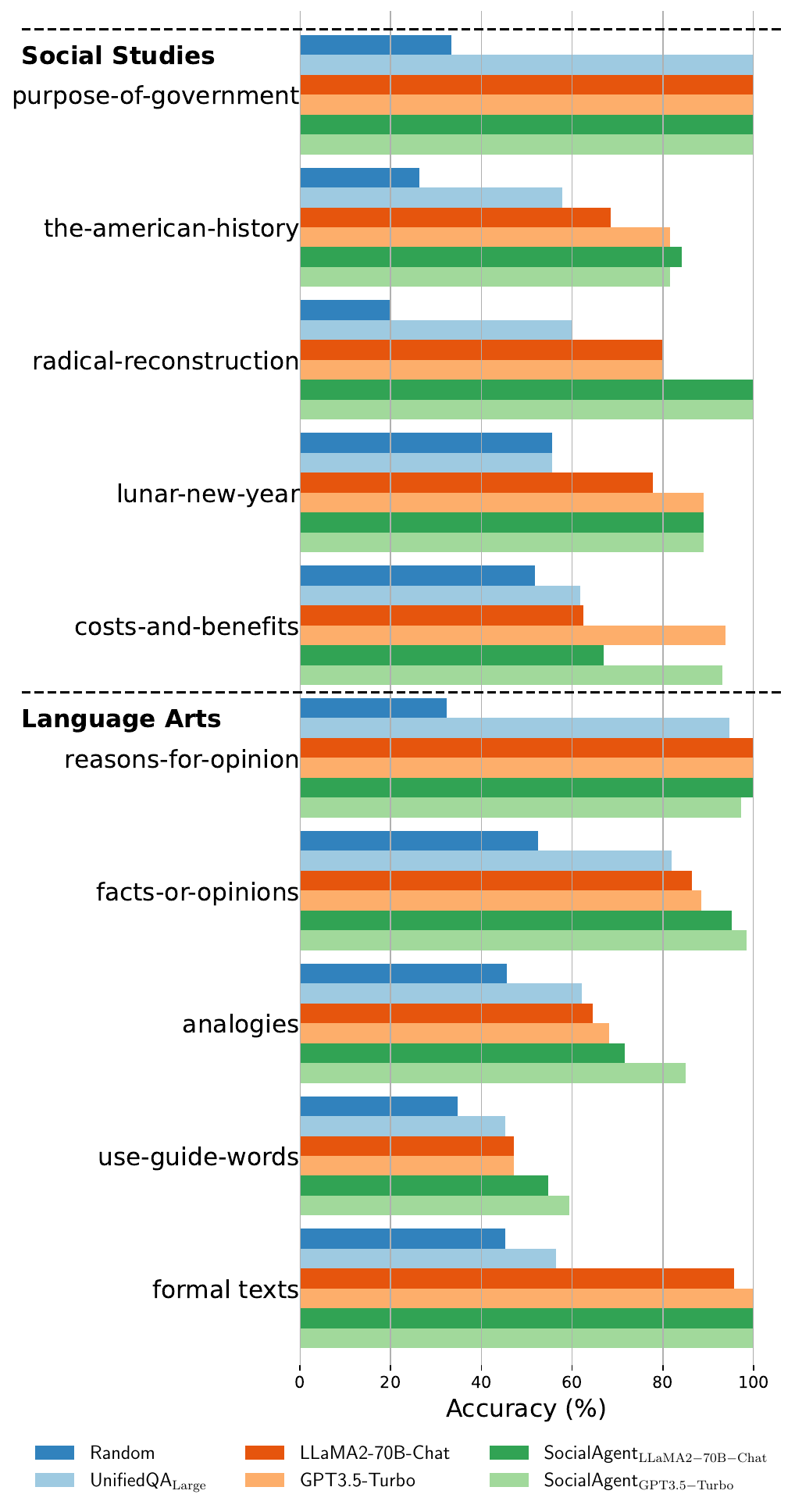}
        \vspace{-0.1in}
\caption{{\small Results on sampled skills of each subject.}}
\label{fig:per_skill}
\end{figure}

\subsection{Main Results}

To examine the social norm understanding, we evaluate models zero-shot on our datasets. The results are shown in Table~\ref{tab:main-exp}. Notably, the recent LLMs such as LLaMA2-70B-Chat and GPT3.5-Turbo improve the performance of previous models by a large margin on average. We also see that GPT3.5-Turbo performs better than LLaMA2-70B-Chat. We observe similar increases in both social studies and language arts. The most improvements (24.4\%) are brought by GPT3.5-Turbo on social studies when compared to the best-performing comparison method UnifiedQA$_{\rm Large}$, and its improvement is 59.7\% over random accuracy. The main reason is that both LLaMA2-70B-Chat and GPT3.5-Turbo are enhanced with reinforcement learning with human feedback (RLHF)~\cite{ouyang2022training}, which is designed to align model responses with human values. This is important to social norm understanding. This is clear from the performance comparison between LLaMA2-70B-Chat and LLaMA2-70B in Figure~\ref{fig:ablation_llama2}. LLaMA2-70B-Chat improves the performance over LLaMA2-70B by 38.6\% on average, and the only notable difference is that LLaMA2-70B-Chat is equipped with the RLHF. This also adds explanations about having more improvements in social studies compared to those in language arts. RLHF mainly brings social perspectives without much emphasis on fundamental language phenomena.

Importantly, \method\ consistently performs the best on both subjects (Table~\ref{tab:main-exp}). With \method, both LLaMA2-70B-Chat and GPT3.5-Turbo improve their performance on our dataset. The best performance is achieved by \method$_{\rm GPT3.5-Turbo}$. This shows that our proposed method is able to integrate important social norm knowledge into LLMs.

\subsection{Results Analysis}
\label{sec:res-analysis}
\paragraph{Skills}
We show the accuracy of comparison models and our method on the skill level in Figure~\ref{fig:per_skill}. We show the results on an uncurated list of skills from both subjects. The complete results on the full skill set are in the appendix. We find similar trends with the overall results on corresponding subjects. Recent advancements in LLMs, in particular, RLHF, improve the performance significantly over the previous near random accuracy. With \method, both GPT3.5-Turbo and LLaMA2-70B-Chat improve their social norm understanding across different social norm skills. GPT3.5-Turbo outperforms LLaMA2-70B-Chat, and \method$_{\rm GPT3.5-Turbo}$ obtains the best overall performance. While there are social norm skills such as ``purpose-of-government'' and ``reasons-for-opinion'' have been mastered by LLMs, there are still plenty of skills such as ``the-american-history'' and ``use-guide-words'' remain unlearned, presenting room for further improvements.

\begin{figure}[ht]
\centering
    \includegraphics[width=1.0\linewidth]{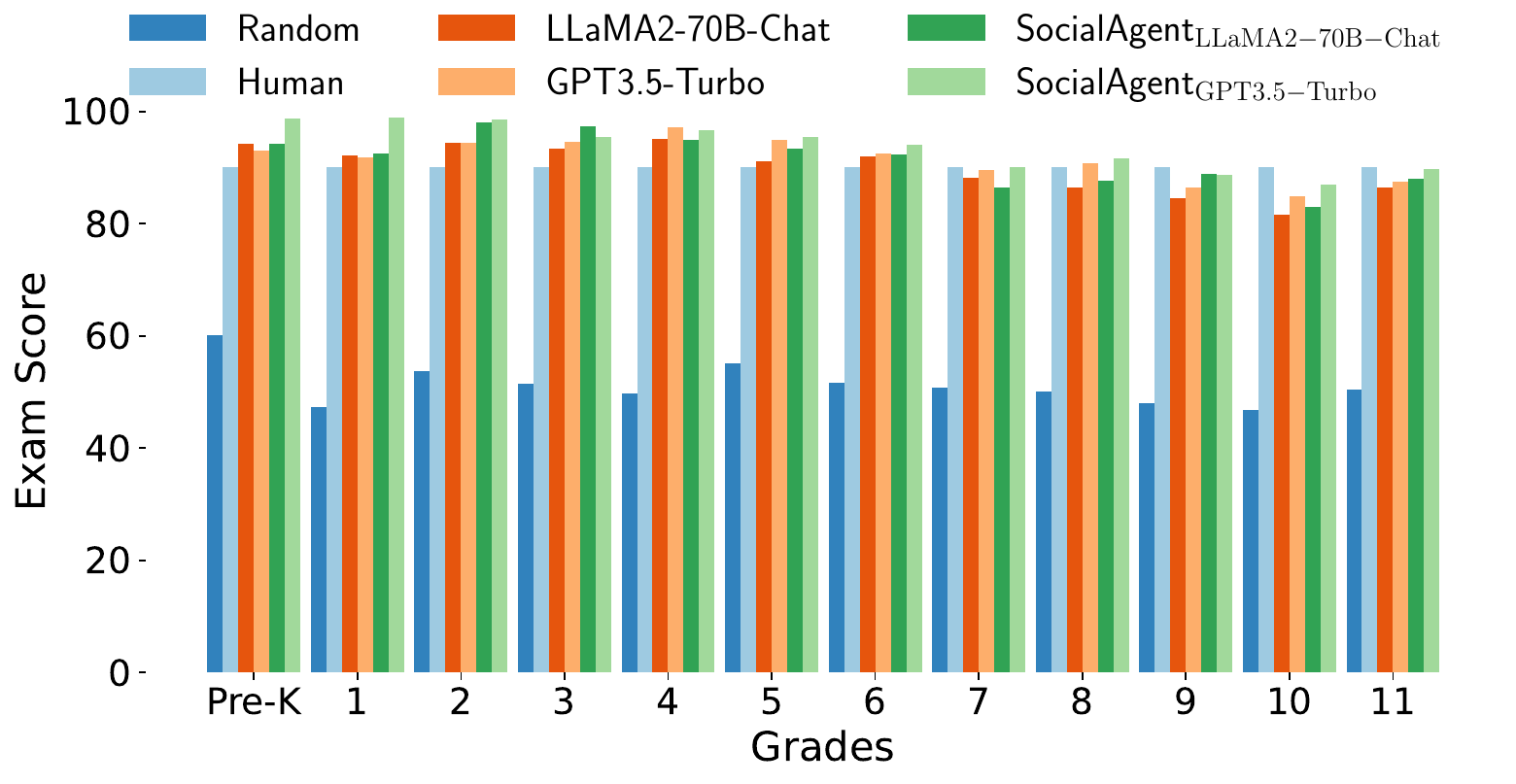}
    \vspace{-0.3in}
    \caption{\small Average grade-level exam scores.}
    \label{fig:grade_smartscore}
\end{figure}

\paragraph{Grades}
Since our dataset provides the fine-grained grade information of the questions, we present the grade-level exam scores of both our method and comparison methods in Figure~\ref{fig:grade_smartscore}. In general, the exam scores of all models decrease slightly when the grades increase. This is because the questions at higher grades are in general more challenging. However, this trend is not obvious based on this human intuition. The reason is that humans learn social norms progressively, while models learn all these skills simultaneously during their training. Besides, we observe similar performance enhancement with previous discussions. For example, recent models improve the performance across all grades significantly compared to the previous random accuracy. \method's brings enhancements in understanding the social norms on all grades. Specifically, all models perform competitively with human performance on lower grades. However, they underperform humans on higher grades such as grades 11 and 12. This indicates significant room to further improve the models' ability to understand advanced social norms.

\paragraph{Scaling Law}
Figure~\ref{fig:ablation_llama2} shows the average accuracy of LLaMA2-Chat of different sizes (7B, 13B, and 70B) with \method. Overall, the model performance increases when the model size gets larger. This indicates that larger models have the additional capacity to learn more accurate social norms during the training. 

\subsection{Ablation Study}
To investigate the importance of each key component of \method\, we show the ablation results of LLaMA2-70B-Chat with \method\ on the social studies subset in Table~\ref{tab:ablation-ma}. The three settings present removing the retrieval agent (``w/o Retrieval Agent''), the programming agent (``w/o Programming Agent''), and the reasoning agent (``w/o Reasoning Agent'') from \method\ respectively. Overall, all components are important since the default \method$_{\rm LLaMA2-70B-Chat}$ obtains the best result. The most significant decrease is brought by removing the reasoning agent. This means that the model has learned certain fundamentals of social norms during their training. So, a better way to prompt the model to obtain the most relevant knowledge is necessary. Moreover, both the retrieval agent and programming agent are essential to incorporate important social norm knowledge into the models.

\begin{table}[ht]
    \centering
    \resizebox{0.75\linewidth}{!}{

\begin{tabular}{@{}ll@{}}
\toprule
Model                        & Social Studies              \\ \midrule
\method$_{\rm LLaMA2-70B-Chat}$                      & \multicolumn{1}{c}{91.8\%} \\
\quad w/o Retrieval Agent    & \multicolumn{1}{c}{91.6\%} \\
\quad w/o Programming Agent  & \multicolumn{1}{c}{91.4\%} \\
\quad w/o Reasoning Agent        & \multicolumn{1}{c}{90.3\%} \\ \bottomrule
\end{tabular}

    }
    \vspace{-0.1in}
    \caption{{\small Ablation results on LLaMA2-70B-Chat with the \method\ method.}}
    \label{tab:ablation-ma}
\end{table}

\subsection{Comparison with Human}
\label{sec:compare_human}
\begin{figure}[!tbh]
\centering

\includegraphics[width=0.9\linewidth]{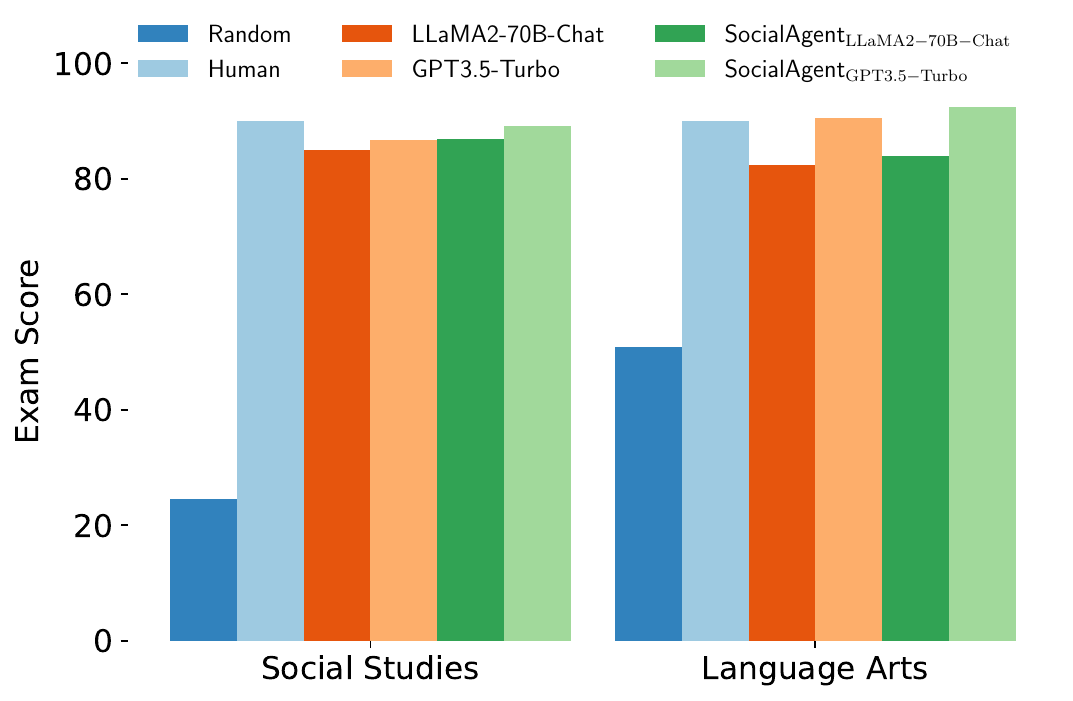}
\vspace{-0.1in}
\caption{\small Compare exam scores of models and humans.}
\label{fig:human_eval}
\vspace{-0.2in}
\end{figure}

It is important to compare models' social norm understanding to that of humans. We compare the exam scores of both GPT3.5-Turbo and LLaMA2-70B-Chat and our methods with millions of elementary student users of the IXL platform. The results are shown in Figure~\ref{fig:human_eval}. Overall, both models still underperform average elementary students in terms of understanding social norms. \method\ helps improve these models to be on par with human performance. For instance, \method$_{\rm GPT3.5-Turbo}$ outperforms humans by an average of 0.8\%. This result is significant although more advancements are needed to compete with human experts.

\subsection{Case Study}
\label{sec:case_study}
\begin{figure}[ht]
    \centering
    \includegraphics[width=1.0\linewidth]{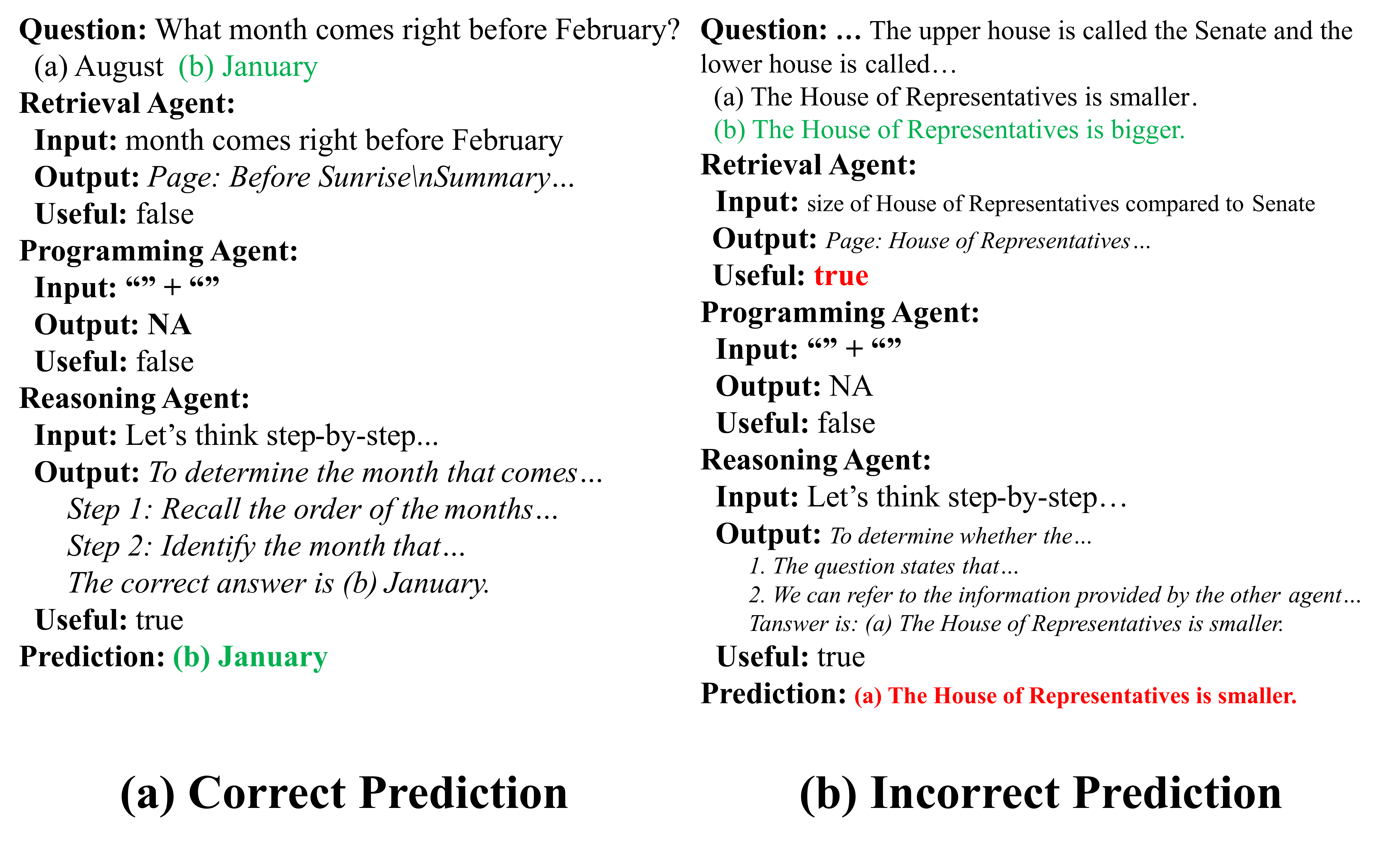}
    \vspace{-0.3in}
    \caption{\small \method$_{\rm GPT3.5-Turbo}$ example predictions.}
    \label{fig:case_study}
    \vspace{-0.2in}
\end{figure}

To better understand what are the models' strengths or weaknesses in understanding social norms, we show examples of best-performing GPT3.5-Turbo predictions with \method. We present an example of correct and incorrect predictions in Figure~\ref{fig:case_study}(a) and (b) respectively. Overall, the models have learned fundamental social norm skills that are concrete and do not require complex reasoning. For example, the answer to the question in Figure~\ref{fig:case_study}(a) is short and relatively straightforward. \method\ is able to utilize the correct model knowledge to deliver the correct answer. Otherwise, the models struggle. For example in Figure~\ref{fig:case_study}(b), the retrieval agent outputs incorrect search results since there is no existing knowledge about this question in the search resources. The other two agent components also do not provide useful context. Based on this, new advancements are needed to help models improve their response quality in challenging scenarios such as long answers and complex reasoning. 

\section{Related Work}

Large language models have demonstrated significant improvements in a variety of NLP tasks recently. LLMs have been introduced and used in real-world applications~\cite{brown2020language,wang-etal-2022-deepstruct,ouyang2022training,openai2023gpt4,chowdhery2023palm,touvron2023llama,taori2023alpaca,vicuna2023,pan2024preparing,wang2023dt}. Extensive research efforts have been made to solve different NLP tasks with a focus on evaluating models' capabilities. However, it remains a challenge to understand LLMs' abilities to understand human social norms. There are existing datasets and benchmarks that aim to help understand the capabilities and limitations of LLMs~\cite{mathdataset,mmlu-stem,scienceqa,liang2022holistic-helm,srivastava2022beyond,shen2024measuring,liu2023fimo,xiong2023trigo}. MMLU~\cite{mmlu-stem} contains 57 tasks spanning broad topics such as maths, science, and history. \citet{scienceqa} collects a multi-choice dataset ScienceQA including social science questions. HELM~\cite{liang2022holistic-helm} is presented to evaluate many aspects of models such as accuracy and robustness on a wide collection of existing tasks such as question answering and toxicity detection. BIG-bench~\cite{srivastava2022beyond} is a benchmark with more than 200 tasks. However, none of these datasets and benchmarks pay attention to evaluation for comprehension of fundamental social norms, which motivates us to present \dataset\ to fill this gap.

There are attempts to connect LLMs with external knowledge, tools and models~\cite{yao2022react,schick2023toolformer,Topsakal_Akinci_2023,liang2023taskmatrixai,wu2023visual,xiong2023dq,wang2020language,shen2022palt,crispino2023agent}. ReAct~\cite{yao2022react} is a general paradigm which combines reasoning and acting with LLMs to solve NLP tasks. \citet{schick2023toolformer} show that LMs can teach themselves to use external tools. LangChain~\cite{Topsakal_Akinci_2023} is a library that aims to benefit the development of LLM based applications. \citet{wu_autogen_2023} propose a multi-agent framework to obtain the answer through the conversations among multiple LLM agents. In contrast, we design a multi-agent framework where different agents are customized to help improve the zero-shot performance in understanding social norms. 

\section{Conclusion}
We introduce a new benchmark for examining LLMs' understanding of social norms. Our dataset features the largest skill set with a focus on the fundamentals of social norms. We evaluated state-of-the-art LLMs including GPT3.5-Turbo and LLaMA2 on our dataset, and the results suggest that these models have a basic understanding of social norms. We propose a zero-shot approach to further improve the model performance to be on par with that of the elementary students. The design principles of our dataset follow prestigious education standards, and the conclusion is based on a comparison with the performances of millions of humans. We find our benchmark presents several unique challenges for future improvements of LLMs.

\section*{Limitations}
For the limitations of our benchmark, the dataset does not contain explanations of the answers yet. Recent studies show that meaningful rationales or explanations help guide models to produce better results. \textcolor{black}{Moreover, the dataset only contains English questions. Therefore language biases or misalignments exist in our dataset.} For our method, the retrieval agent uses Wikipedia as the only source. Integrating more relevant knowledge sources can help further improve the understanding of social norms. Finally, our method shares some common limitations with most deep learning approaches. For example, the decisions are not easy to interpret.

\section*{Ethical Impact}
We hereby acknowledge that all of the co-authors of this work are aware of the provided ACL Code of Ethics and honor the code of conduct. 
We collect data from online sources, which do not contain any personal information or inappropriate content that may cause negative societal impacts. We cited the data creators and the copyright belongs to the original data owners. The \dataset\ dataset is under the CC BY-NC-SA 4.0 license (Creative Commons Attribution-NonCommercial-ShareAlike 4.0 International) and is used for non-commercial research purposes. We evaluate and develop methods based on large language models. The potential risks of using such models are discussed in the original papers~\cite{ouyang2022training,unifiedqa,touvron2023llama-2}.

\section*{Acknowledgements}
This paper is partially supported by the National Natural Science Foundation of China with Grant No.62276002.

\bibliography{anthology,custom}

\begin{thebibliography}{49}
\expandafter\ifx\csname natexlab\endcsname\relax\def\natexlab#1{#1}\fi

\bibitem[{Bashkov et~al.(2021)Bashkov, Mattison, and Hochstein}]{bashkov2021ixl}
Bozhidar~M Bashkov, Kate Mattison, and Lara Hochstein. 2021.
\newblock Ixl design principles.

\bibitem[{Brown et~al.(2020)Brown, Mann, Ryder, Subbiah et~al.}]{brown2020language}
Tom Brown, Benjamin Mann, Nick Ryder, Melanie Subbiah, et~al. 2020.
\newblock Language models are few-shot learners.
\newblock \emph{Advances in neural information processing systems}, 33:1877--1901.

\bibitem[{Chen et~al.(2023)Chen, Zaharia, and Zou}]{chen2023chatgpts}
Lingjiao Chen, Matei Zaharia, and James Zou. 2023.
\newblock \href {http://arxiv.org/abs/2307.09009} {How is chatgpt's behavior changing over time?}

\bibitem[{Chiang et~al.(2023)Chiang, Li, Lin, Sheng, and others.}]{vicuna2023}
Wei-Lin Chiang, Zhuohan Li, Zi~Lin, Ying Sheng, and others. 2023.
\newblock \href {https://lmsys.org/blog/2023-03-30-vicuna/} {Vicuna: An open-source chatbot impressing gpt-4 with 90\%* chatgpt quality}.

\bibitem[{Chowdhery et~al.(2023)Chowdhery, Narang, Devlin, Bosma et~al.}]{chowdhery2023palm}
Aakanksha Chowdhery, Sharan Narang, Jacob Devlin, Maarten Bosma, et~al. 2023.
\newblock Palm: Scaling language modeling with pathways.
\newblock \emph{Journal of Machine Learning Research}, 24(240):1--113.

\bibitem[{Crispino et~al.(2023)Crispino, Montgomery, Zeng, Song, and Wang}]{crispino2023agent}
Nicholas Crispino, Kyle Montgomery, Fankun Zeng, Dawn Song, and Chenguang Wang. 2023.
\newblock Agent instructs large language models to be general zero-shot reasoners.
\newblock \emph{arXiv preprint arXiv:2310.03710}.

\bibitem[{Gemini~Team(2023)}]{gemini}
Google Gemini~Team. 2023.
\newblock Gemini: A family of highly capable multimodal models.
\newblock \url{https://storage.googleapis.com/deepmind-media/gemini/gemini_1_report.pdf}.

\bibitem[{Guo et~al.(2017)Guo, Pleiss, Sun, and Weinberger}]{calibration}
Chuan Guo, Geoff Pleiss, Yu~Sun, and Kilian~Q. Weinberger. 2017.
\newblock On calibration of modern neural networks.
\newblock In \emph{ICML}, pages 1321--1330.

\bibitem[{Hendrycks et~al.(2021{\natexlab{a}})Hendrycks, Burns, Basart, Zou, Mazeika, Song, and Steinhardt}]{mmlu-stem}
Dan Hendrycks, Collin Burns, Steven Basart, Andy Zou, Mantas Mazeika, Dawn Song, and Jacob Steinhardt. 2021{\natexlab{a}}.
\newblock Measuring massive multitask language understanding.
\newblock In \emph{ICLR}.

\bibitem[{Hendrycks et~al.(2021{\natexlab{b}})Hendrycks, Burns, Kadavath, Arora, Basart, Tang, Song, and Steinhardt}]{mathdataset}
Dan Hendrycks, Collin Burns, Saurav Kadavath, Akul Arora, Steven Basart, Eric Tang, Dawn Song, and Jacob Steinhardt. 2021{\natexlab{b}}.
\newblock Measuring mathematical problem solving with the math dataset.
\newblock \emph{NeurIPS}.

\bibitem[{Hochreiter and Schmidhuber(1997)}]{hochreiter1997long-lstm}
Sepp Hochreiter and J{\"u}rgen Schmidhuber. 1997.
\newblock Long short-term memory.
\newblock \emph{Neural computation}, 9(8):1735--1780.

\bibitem[{IXL(2014{\natexlab{a}})}]{IXLwork}
IXL. 2014{\natexlab{a}}.
\newblock How does the smartscore work?
\newblock \url{https://www.ixl.com/help-center/article/1272663/how_does_the_smartscore_work}.

\bibitem[{IXL(2014{\natexlab{b}})}]{IXLGuide}
IXL. 2014{\natexlab{b}}.
\newblock Understanding the ixl smartscore.
\newblock \url{https://blog.ixl.com/wp-content/uploads/2014/11/SmartScore-guide.pdf}.

\bibitem[{Khashabi et~al.(2020)Khashabi, Min, Khot, Sabharwal, Tafjord, Clark, and Hajishirzi}]{unifiedqa}
Daniel Khashabi, Sewon Min, Tushar Khot, Ashish Sabharwal, Oyvind Tafjord, Peter Clark, and Hannaneh Hajishirzi. 2020.
\newblock Unifiedqa: Crossing format boundaries with a single {QA} system.
\newblock In \emph{Findings of EMNLP}, pages 1896--1907.

\bibitem[{Learning(2019)}]{learning2019impact}
IXL Learning. 2019.
\newblock The impact of ixl math and ixl ela on student achievement in grades pre-k to 12.

\bibitem[{Li et~al.(2023)Li, Hammoud, Itani, Khizbullin, and Ghanem}]{li2023camel}
Guohao Li, Hasan Abed Al~Kader Hammoud, Hani Itani, Dmitrii Khizbullin, and Bernard Ghanem. 2023.
\newblock Camel: Communicative agents for" mind" exploration of large scale language model society.
\newblock \emph{arXiv preprint arXiv:2303.17760}.

\bibitem[{Liang et~al.(2022)Liang, Bommasani, Lee et~al.}]{liang2022holistic-helm}
Percy Liang, Rishi Bommasani, Tony Lee, et~al. 2022.
\newblock Holistic evaluation of language models.
\newblock \emph{arXiv preprint arXiv:2211.09110}.

\bibitem[{Liang et~al.(2023)Liang, Wu, Song et~al.}]{liang2023taskmatrixai}
Yaobo Liang, Chenfei Wu, Ting Song, et~al. 2023.
\newblock \href {http://arxiv.org/abs/2303.16434} {Taskmatrix.ai: Completing tasks by connecting foundation models with millions of apis}.

\bibitem[{Liu et~al.(2023{\natexlab{a}})Liu, Shen et~al.}]{liu2023fimo}
Chengwu Liu, Jianhao Shen, et~al. 2023{\natexlab{a}}.
\newblock Fimo: A challenge formal dataset for automated theorem proving.
\newblock \emph{arXiv preprint arXiv:2309.04295}.

\bibitem[{Liu et~al.(2023{\natexlab{b}})Liu, Yao, Zhang, Xue, Heinecke, Murthy, Feng, Chen, Niebles, Arpit et~al.}]{liu2023bolaa}
Zhiwei Liu, Weiran Yao, Jianguo Zhang, Le~Xue, Shelby Heinecke, Rithesh Murthy, Yihao Feng, Zeyuan Chen, Juan~Carlos Niebles, Devansh Arpit, et~al. 2023{\natexlab{b}}.
\newblock Bolaa: Benchmarking and orchestrating llm-augmented autonomous agents.
\newblock \emph{arXiv preprint arXiv:2308.05960}.

\bibitem[{Lu et~al.(2022)Lu, Mishra, Xia, Qiu, Chang, Zhu, Tafjord, Clark, and Kalyan}]{scienceqa}
Pan Lu, Swaroop Mishra, Tony Xia, Liang Qiu, Kai-Wei Chang, Song-Chun Zhu, Oyvind Tafjord, Peter Clark, and Ashwin Kalyan. 2022.
\newblock Learn to explain: Multimodal reasoning via thought chains for science question answering.
\newblock In \emph{NeurIPS}.

\bibitem[{OpenAI(2023)}]{openai2023gpt4}
OpenAI. 2023.
\newblock \href {http://arxiv.org/abs/2303.08774} {Gpt-4 technical report}.

\bibitem[{Ouyang et~al.(2022)Ouyang, Wu, Jiang, Almeida, Wainwright, Mishkin, Zhang, Agarwal, Slama, Ray et~al.}]{ouyang2022training}
Long Ouyang, Jeffrey Wu, Xu~Jiang, Diogo Almeida, Carroll Wainwright, Pamela Mishkin, Chong Zhang, Sandhini Agarwal, Katarina Slama, Alex Ray, et~al. 2022.
\newblock Training language models to follow instructions with human feedback.
\newblock \emph{Advances in neural information processing systems}, 35:27730--27744.

\bibitem[{Pan et~al.(2019)Pan, Xu, Wang, Ye, Wang, Bai, and Xu}]{DBLP:conf/aaai/PanXWYWBX19}
Yu~Pan, Jing Xu, Maolin Wang, Jinmian Ye, Fei Wang, Kun Bai, and Zenglin Xu. 2019.
\newblock Compressing recurrent neural networks with tensor ring for action recognition.
\newblock In \emph{{AAAI}}, pages 4683--4690. {AAAI} Press.

\bibitem[{Pan et~al.(2024)Pan, Yuan, Yin, Shi, Xu, Zhang, Shang, Jiang, and Liu}]{pan2024preparing}
Yu~Pan, Ye~Yuan, Yichun Yin, Jiaxin Shi, Zenglin Xu, Ming Zhang, Lifeng Shang, Xin Jiang, and Qun Liu. 2024.
\newblock Preparing lessons for progressive training on language models.
\newblock \emph{arXiv preprint arXiv:2401.09192}.

\bibitem[{Raffel et~al.(2020)Raffel, Shazeer, Roberts, Lee, Narang, Matena, Zhou, Li, and Liu}]{raffel2020exploring-t5}
Colin Raffel, Noam Shazeer, Adam Roberts, Katherine Lee, Sharan Narang, Michael Matena, Yanqi Zhou, Wei Li, and Peter~J Liu. 2020.
\newblock Exploring the limits of transfer learning with a unified text-to-text transformer.
\newblock \emph{The Journal of Machine Learning Research}, 21(1):5485--5551.

\bibitem[{Rumelhart et~al.(1986)Rumelhart, Hinton, and Williams}]{rumelhart1986learning-rnn}
David~E Rumelhart, Geoffrey~E Hinton, and Ronald~J Williams. 1986.
\newblock Learning internal representations by error propagation, parallel distributed processing, explorations in the microstructure of cognition, ed. de rumelhart and j. mcclelland. vol. 1. 1986.
\newblock \emph{Biometrika}, 71:599--607.

\bibitem[{Schick et~al.(2023)Schick, Dwivedi-Yu, Dessì, Raileanu, Lomeli, Zettlemoyer, Cancedda, and Scialom}]{schick2023toolformer}
Timo Schick, Jane Dwivedi-Yu, Roberto Dessì, Roberta Raileanu, Maria Lomeli, Luke Zettlemoyer, Nicola Cancedda, and Thomas Scialom. 2023.
\newblock \href {http://arxiv.org/abs/2302.04761} {Toolformer: Language models can teach themselves to use tools}.

\bibitem[{Shen et~al.(2022)Shen, Wang, Yuan et~al.}]{shen2022palt}
Jianhao Shen, Chenguang Wang, Ye~Yuan, et~al. 2022.
\newblock Palt: parameter-lite transfer of language models for knowledge graph completion.
\newblock \emph{arXiv preprint arXiv:2210.13715}.

\bibitem[{Shen et~al.(2024)Shen, Yuan, Mirzoyan, Zhang, and Wang}]{shen2024measuring}
Jianhao Shen, Ye~Yuan, Srbuhi Mirzoyan, Ming Zhang, and Chenguang Wang. 2024.
\newblock Measuring vision-language stem skills of neural models.
\newblock \emph{arXiv preprint arXiv:2402.17205}.

\bibitem[{Shi et~al.(2023)Shi, Chen, Misra, Scales, Dohan, Chi, Sch{\"a}rli, and Zhou}]{shi2023large}
Freda Shi, Xinyun Chen, Kanishka Misra, Nathan Scales, David Dohan, Ed~H Chi, Nathanael Sch{\"a}rli, and Denny Zhou. 2023.
\newblock Large language models can be easily distracted by irrelevant context.
\newblock In \emph{International Conference on Machine Learning}, pages 31210--31227. PMLR.

\bibitem[{Srivastava et~al.(2022)Srivastava, Rastogi, Rao, Shoeb, Abid, Fisch, Brown, Santoro, Gupta, Garriga-Alonso et~al.}]{srivastava2022beyond}
Aarohi Srivastava, Abhinav Rastogi, Abhishek Rao, Abu Awal~Md Shoeb, Abubakar Abid, Adam Fisch, Adam~R Brown, Adam Santoro, Aditya Gupta, Adri{\`a} Garriga-Alonso, et~al. 2022.
\newblock Beyond the imitation game: Quantifying and extrapolating the capabilities of language models.
\newblock \emph{arXiv preprint arXiv:2206.04615}.

\bibitem[{Taori et~al.(2023)Taori, Gulrajani, Zhang, Dubois, Li, Guestrin, Liang, and Hashimoto}]{taori2023alpaca}
Rohan Taori, Ishaan Gulrajani, Tianyi Zhang, Yann Dubois, Xuechen Li, Carlos Guestrin, Percy Liang, and Tatsunori~B Hashimoto. 2023.
\newblock Alpaca: A strong, replicable instruction-following model.
\newblock \emph{Stanford Center for Research on Foundation Models. https://crfm. stanford. edu/2023/03/13/alpaca. html}, 3(6):7.

\bibitem[{Topsakal and Akinci(2023)}]{Topsakal_Akinci_2023}
Oguzhan Topsakal and Tahir~Cetin Akinci. 2023.
\newblock \href {https://doi.org/10.59287/icaens.1127} {Creating large language model applications utilizing langchain: A primer on developing llm apps fast}.
\newblock \emph{International Conference on Applied Engineering and Natural Sciences}, 1(1):1050–1056.

\bibitem[{Touvron et~al.(2023{\natexlab{a}})Touvron, Lavril, Izacard, Martinet, Lachaux, Lacroix, Rozière, Goyal, Hambro, Azhar, Rodriguez, Joulin, Grave, and Lample}]{touvron2023llama}
Hugo Touvron, Thibaut Lavril, Gautier Izacard, Xavier Martinet, Marie-Anne Lachaux, Timothée Lacroix, Baptiste Rozière, Naman Goyal, Eric Hambro, Faisal Azhar, Aurelien Rodriguez, Armand Joulin, Edouard Grave, and Guillaume Lample. 2023{\natexlab{a}}.
\newblock \href {http://arxiv.org/abs/2302.13971} {Llama: Open and efficient foundation language models}.

\bibitem[{Touvron et~al.(2023{\natexlab{b}})Touvron, Martin, Stone, Albert, Almahairi, Babaei, Bashlykov, Batra, Bhargava, Bhosale et~al.}]{touvron2023llama-2}
Hugo Touvron, Louis Martin, Kevin Stone, Peter Albert, Amjad Almahairi, Yasmine Babaei, Nikolay Bashlykov, Soumya Batra, Prajjwal Bhargava, Shruti Bhosale, et~al. 2023{\natexlab{b}}.
\newblock Llama 2: Open foundation and fine-tuned chat models.
\newblock \emph{arXiv preprint arXiv:2307.09288}.

\bibitem[{Tu et~al.(2023)Tu, Li, Yu, Wang, Hou, and Li}]{tu2023chatlog}
Shangqing Tu, Chunyang Li, Jifan Yu, Xiaozhi Wang, Lei Hou, and Juanzi Li. 2023.
\newblock Chatlog: Recording and analyzing chatgpt across time.
\newblock \emph{arXiv preprint arXiv:2304.14106}.

\bibitem[{Wang et~al.(2022)Wang, Liu, Chen, Hong, Tang, and Song}]{wang-etal-2022-deepstruct}
Chenguang Wang, Xiao Liu, Zui Chen, Haoyun Hong, Jie Tang, and Dawn Song. 2022.
\newblock \href {https://doi.org/10.18653/v1/2022.findings-acl.67} {{D}eep{S}truct: Pretraining of language models for structure prediction}.
\newblock In \emph{Findings of the Association for Computational Linguistics: ACL 2022}, pages 803--823, Dublin, Ireland. Association for Computational Linguistics.

\bibitem[{Wang et~al.(2020)Wang, Liu, and Song}]{wang2020language}
Chenguang Wang, Xiao Liu, and Dawn Song. 2020.
\newblock Language models are open knowledge graphs.
\newblock \emph{arXiv preprint arXiv:2010.11967}.

\bibitem[{Wang et~al.(2023{\natexlab{a}})Wang, Yuan, Liu, Shen, Yin, Xiong, Xie, Shi, Li, Li et~al.}]{wang2023dt}
Haiming Wang, Ye~Yuan, Zhengying Liu, Jianhao Shen, Yichun Yin, Jing Xiong, Enze Xie, Han Shi, Yujun Li, Lin Li, et~al. 2023{\natexlab{a}}.
\newblock Dt-solver: Automated theorem proving with dynamic-tree sampling guided by proof-level value function.
\newblock In \emph{Proceedings of the 61st Annual Meeting of the Association for Computational Linguistics (Volume 1: Long Papers)}, pages 12632--12646.

\bibitem[{Wang et~al.(2023{\natexlab{b}})Wang, Pan, Xu, Yang, Li, and Cichocki}]{wang2023tensor}
Maolin Wang, Yu~Pan, Zenglin Xu, Xiangli Yang, Guangxi Li, and Andrzej Cichocki. 2023{\natexlab{b}}.
\newblock Tensor networks meet neural networks: A survey and future perspectives.
\newblock \emph{arXiv preprint arXiv:2302.09019}.

\bibitem[{Wei et~al.(2022)Wei, Wang, Schuurmans, Bosma, Xia, Chi, Le, Zhou et~al.}]{wei2022chain-cot}
Jason Wei, Xuezhi Wang, Dale Schuurmans, Maarten Bosma, Fei Xia, Ed~Chi, Quoc~V Le, Denny Zhou, et~al. 2022.
\newblock Chain-of-thought prompting elicits reasoning in large language models.
\newblock \emph{Advances in Neural Information Processing Systems}, 35:24824--24837.

\bibitem[{Wu et~al.(2023{\natexlab{a}})Wu, Yin, Qi, Wang, Tang, and Duan}]{wu2023visual}
Chenfei Wu, Shengming Yin, Weizhen Qi, Xiaodong Wang, Zecheng Tang, and Nan Duan. 2023{\natexlab{a}}.
\newblock \href {http://arxiv.org/abs/2303.04671} {Visual chatgpt: Talking, drawing and editing with visual foundation models}.

\bibitem[{Wu et~al.(2023{\natexlab{b}})Wu, Bansal, Zhang, Wu, Zhang, Zhu, Li, Jiang, Zhang, and Wang}]{wu_autogen_2023}
Qingyun Wu, Gagan Bansal, Jieyu Zhang, Yiran Wu, Shaokun Zhang, Erkang Zhu, Beibin Li, Li~Jiang, Xiaoyun Zhang, and Chi Wang. 2023{\natexlab{b}}.
\newblock \href {https://doi.org/10.48550/arXiv.2308.08155} {{AutoGen}: {Enabling} {Next}-{Gen} {LLM} {Applications} via {Multi}-{Agent} {Conversation} {Framework}}.
\newblock ArXiv:2308.08155 [cs].

\bibitem[{Xi et~al.(2023)Xi, Chen, Guo, He, Ding, Hong, Zhang, Wang, Jin, Zhou et~al.}]{xi2023agent-survey}
Zhiheng Xi, Wenxiang Chen, Xin Guo, Wei He, Yiwen Ding, Boyang Hong, Ming Zhang, Junzhe Wang, Senjie Jin, Enyu Zhou, et~al. 2023.
\newblock The rise and potential of large language model based agents: A survey.
\newblock \emph{arXiv preprint arXiv:2309.07864}.

\bibitem[{Xiong et~al.(2023{\natexlab{a}})Xiong, Li, Zheng, Guo, Yin, Xie, Yang, Cao, Wang, Han et~al.}]{xiong2023dq}
Jing Xiong, Zixuan Li, Chuanyang Zheng, Zhijiang Guo, Yichun Yin, Enze Xie, Zhicheng Yang, Qingxing Cao, Haiming Wang, Xiongwei Han, et~al. 2023{\natexlab{a}}.
\newblock Dq-lore: Dual queries with low rank approximation re-ranking for in-context learning.
\newblock \emph{arXiv preprint arXiv:2310.02954}.

\bibitem[{Xiong et~al.(2023{\natexlab{b}})Xiong, Shen, Yuan, Wang, Yin, Liu, Li, Guo, Cao, Huang et~al.}]{xiong2023trigo}
Jing Xiong, Jianhao Shen, Ye~Yuan, Haiming Wang, Yichun Yin, Zhengying Liu, Lin Li, Zhijiang Guo, Qingxing Cao, Yinya Huang, et~al. 2023{\natexlab{b}}.
\newblock Trigo: Benchmarking formal mathematical proof reduction for generative language models.
\newblock \emph{arXiv preprint arXiv:2310.10180}.

\bibitem[{Yao et~al.(2022)Yao, Zhao, Yu, Du, Shafran, Narasimhan, and Cao}]{yao2022react}
Shunyu Yao, Jeffrey Zhao, Dian Yu, Nan Du, Izhak Shafran, Karthik Narasimhan, and Yuan Cao. 2022.
\newblock React: Synergizing reasoning and acting in language models.
\newblock \emph{arXiv preprint arXiv:2210.03629}.

\bibitem[{Yin et~al.(2023)Yin, Sun et~al.}]{yin2023exchangeofthought}
Zhangyue Yin, Qiushi Sun, et~al. 2023.
\newblock \href {http://arxiv.org/abs/2312.01823} {Exchange-of-thought: Enhancing large language model capabilities through cross-model communication}.

\end{thebibliography}
\appendix
\clearpage

\section{More Details on \dataset }
\label{apped:data_detail}
In this section, we provide more details on \dataset, including dataset analysis, models, and dataset collection.

\subsection{Analysis}
\begin{figure}[ht]
    \centering
    \includegraphics[width=\linewidth]{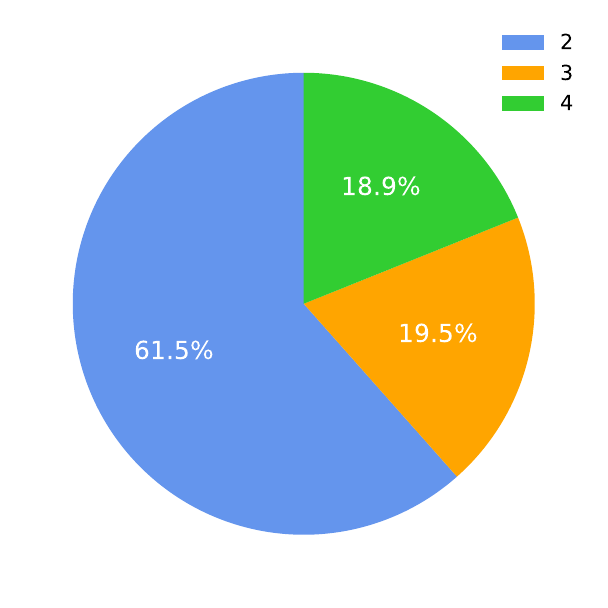}
    \caption{{\small \#Answers distribution.}}
    \label{fig:num_ans_dist}
\end{figure}

\begin{figure}[ht]
    \centering
    \includegraphics[width=\linewidth]{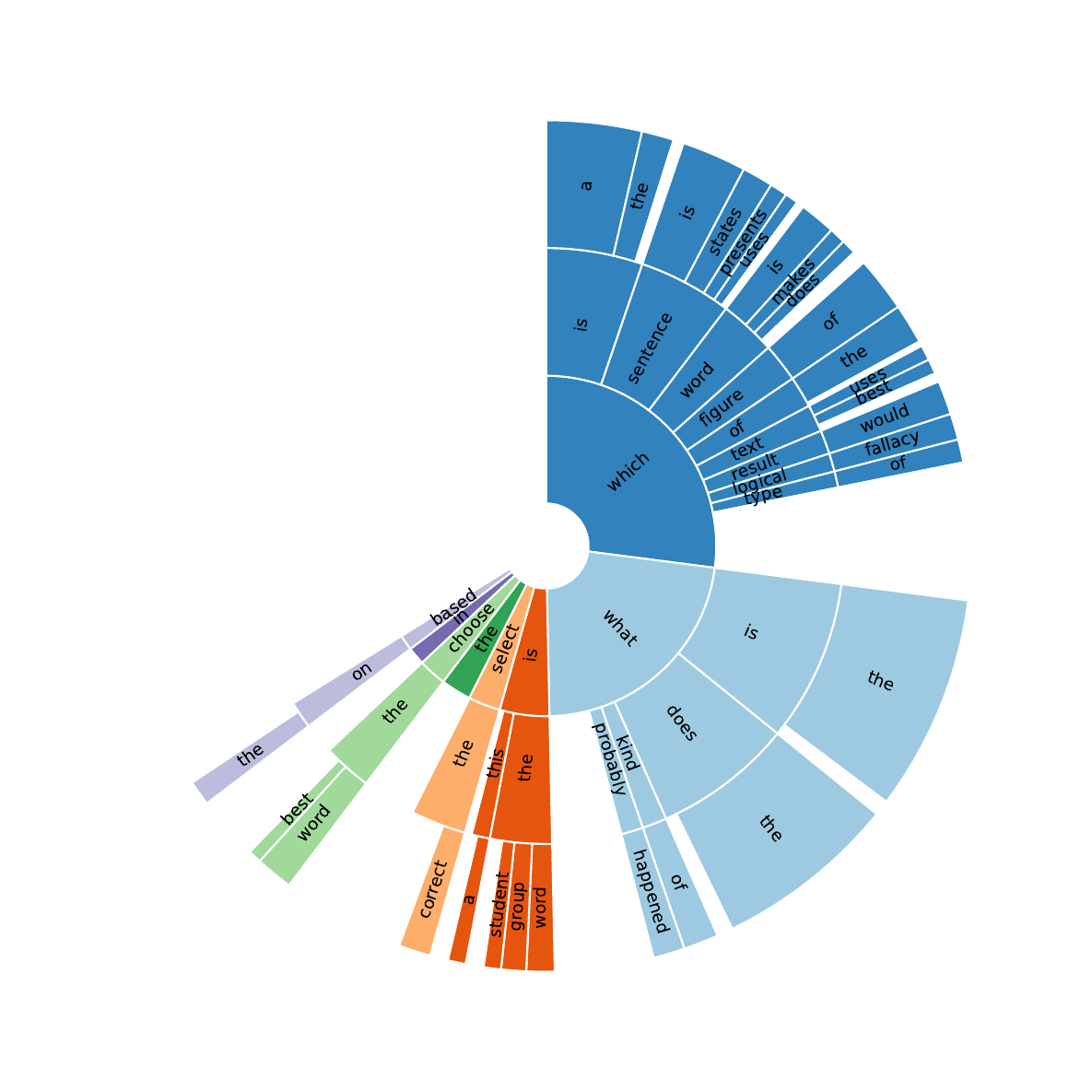}
    \caption{{\small Question type distribution.}}
    \label{fig:trigram}
\end{figure}

\begin{figure}[ht]
        \centering
    \includegraphics[width=\linewidth]{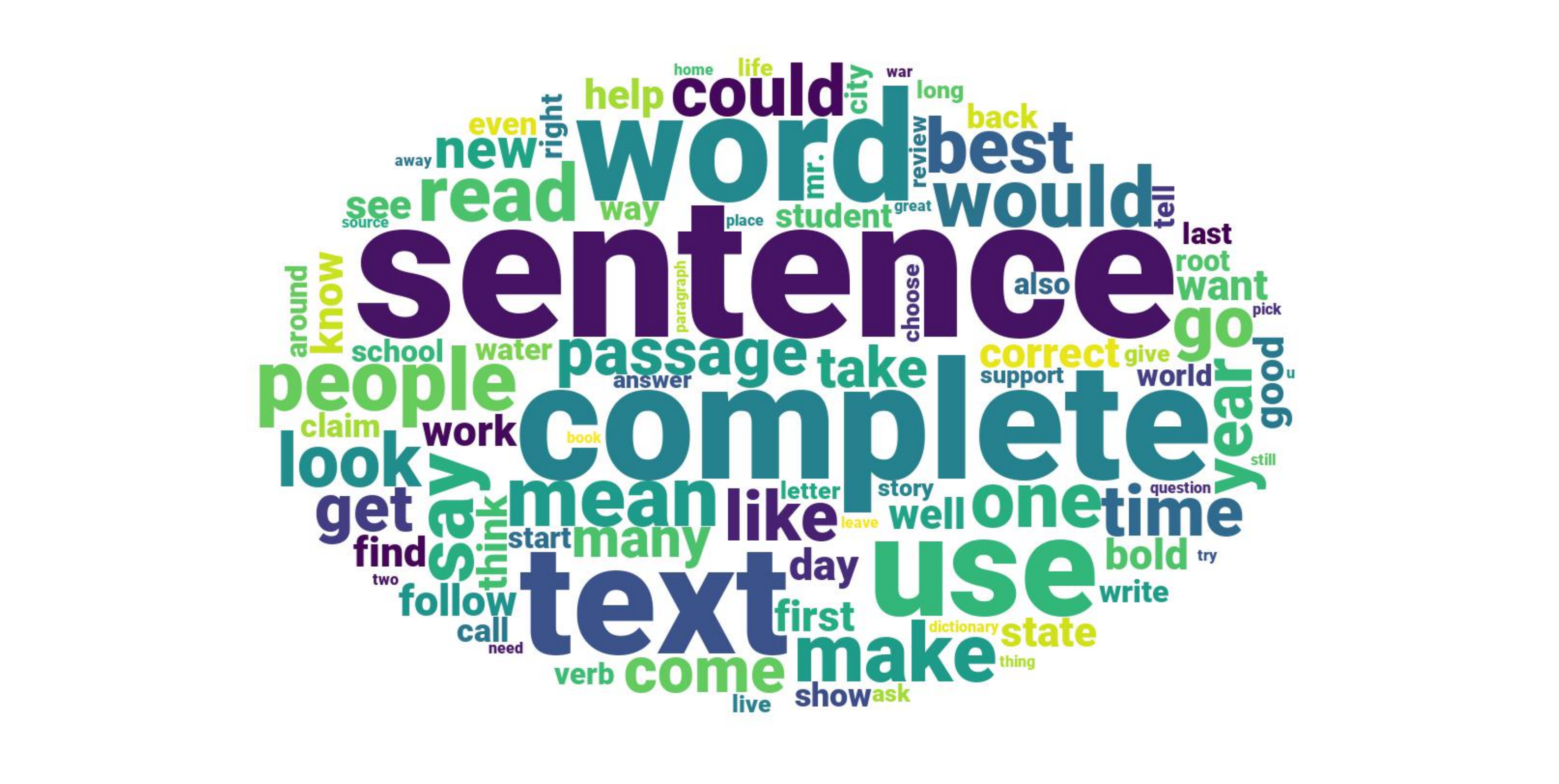}
    \caption{\small Word cloud of question texts in \dataset.}
    \label{fig:word_cloud}
\end{figure}

\begin{figure}[ht]
    \centering
    \includegraphics[width=\linewidth]{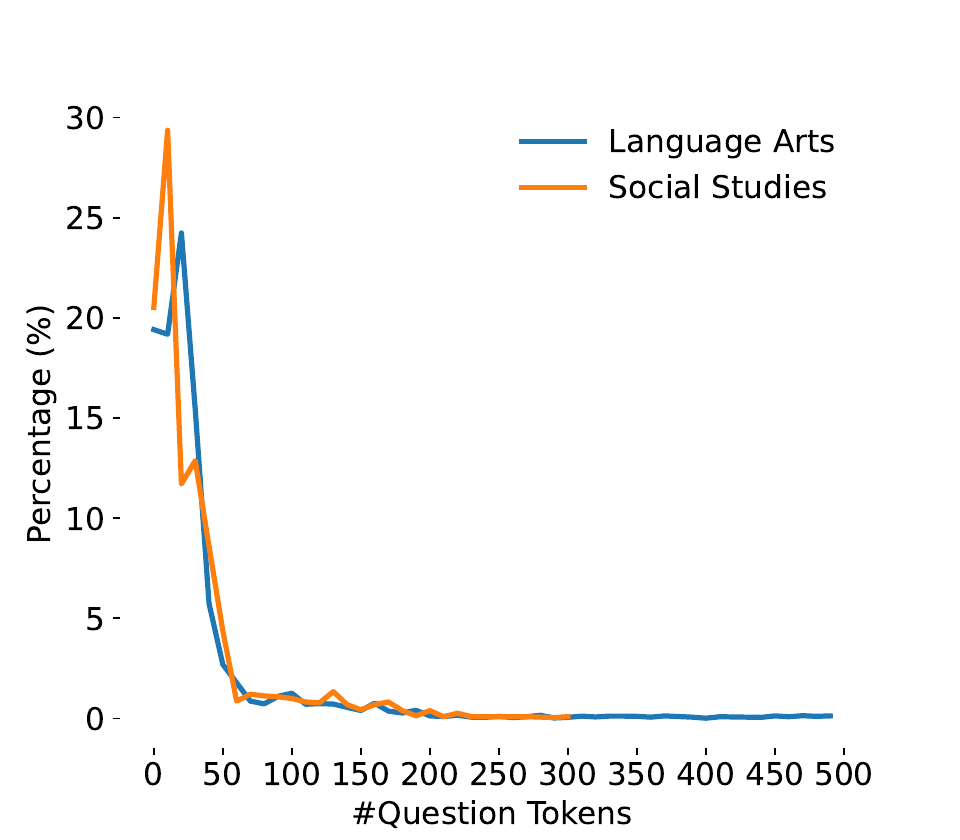}
    \caption{{\small Question length distribution.}}
    \label{fig:ques_len_dist}    
\end{figure}

\begin{figure}[ht]
    \centering
    \includegraphics[width=\linewidth]{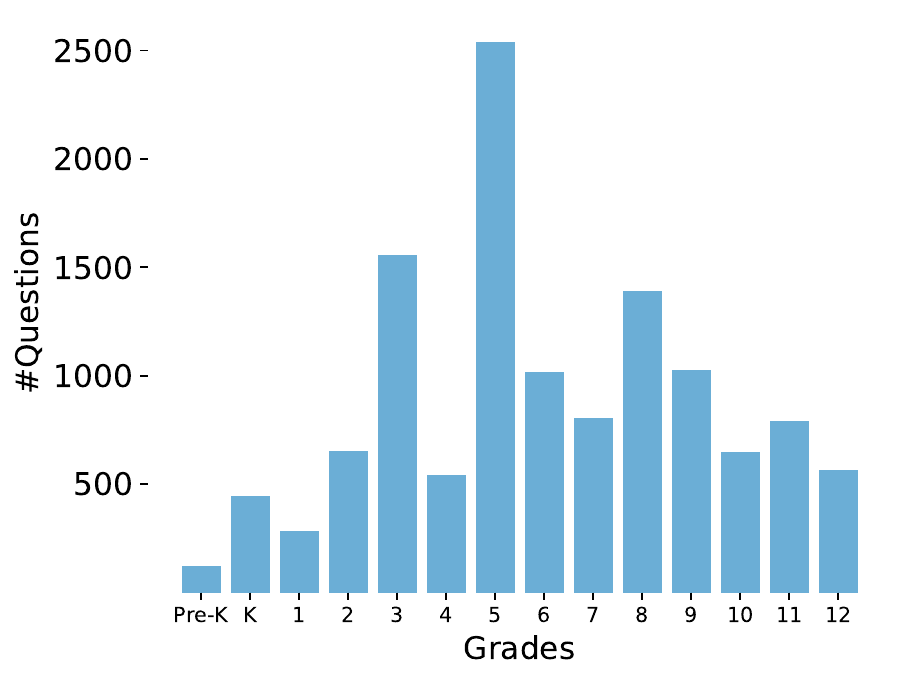}
    \caption{{\small \#Questions per grade.}}
    \label{fig:ques_per_grade}
\end{figure}

\paragraph{Questions and Answers} \dataset\ contains multi-choice questions (Appendix~\ref{apped:summary_skills} provides a question example for each skill). The question contains a textual description with an optional textual context. We further analyze the questions from the following aspects. 
(\expandafter{\romannumeral1}) The number of answers. \dataset\ has averaging $2.6$ answer options for each question. The distribution is presented in Figure~\ref{fig:num_ans_dist}. In practice, the more answer options one question has, the more difficult it is. 
(\expandafter{\romannumeral2}) Question type. We categorize questions based on the first three words of the question text as shown in Figure \ref{fig:trigram}. \dataset\ mostly includes factoid questions that start with words such as ``which'' and ``what''. We also show the word cloud of our \dataset\ in Figure \ref{fig:word_cloud}. We can see the most common words like ``sentence'' and ``complete''. This indicates that many questions are sentence-completion type.
(\expandafter{\romannumeral3}) Question distribution. Figure~\ref{fig:ques_len_dist} depicts the distribution of question lengths. We can see both subjects generally follow a long-tail distribution, while language arts distribution has a longer tail because it includes many long reading comprehension questions. Heuristically, longer questions are more difficult to solve. Figure ~\ref{fig:ques_per_grade} shows the number of questions in each grade. The questions are primarily distributed between grades 3 and 9, accounting for 72\% of the total.

\subsection{Dataset Collection}
We collect the Social Studies and Language Arts datasets from \textit{IXL}\footnote{\url{https://www.ixl.com/}}. We collect all the multi-choice questions which only contain texts. All questions have only one correct answer. We collect 200 trials for each skill and remove the duplicated problems. Finally, Language Arts problems are much more than Social Studies, so we sample problems from each skill of Language Arts subject uniformly.

\subsection{Additional Related Work}
\textcolor{black}{There are existing agent frameworks, such as ReAct~\citep{yao2022react}, Exchange-of-Thought~\citep{yin2023exchangeofthought}, BOLAA~\citep{liu2023bolaa}, etc. ReAct needs multiple turns of dialog, which often leads the models to forget the long dependency knowledge. Exchange-of-Thought uses multiple turns of dialog with different characters, and each character conducts its analysis of the question independently. Instead, each module in our framework focuses on its specific functionality, and the agent decides which output (or combinations) to use on the fly. BOLAA mainly enables LLM to do planning when solving problems. We find that planning is not always useful when answering social norms questions, and the planning ability of LLaMA-2 and GPT-3.5-Turbo are limited and present additional risks such as misleading the models.}

\section{More Details on Experiments}
\subsection{Experimental Setup}
\begin{figure*}[!t]
\centering
\includegraphics[width=0.96\linewidth]{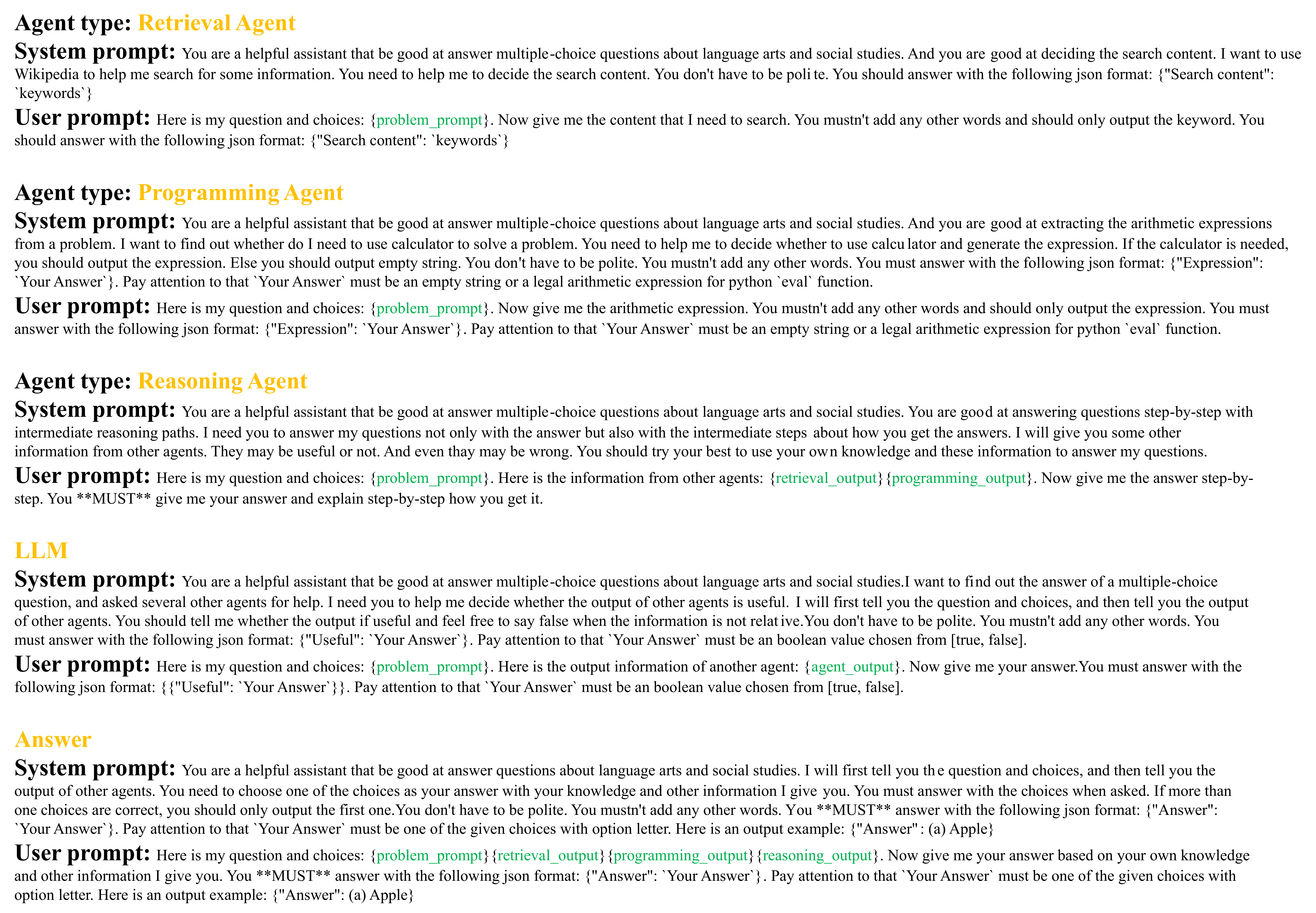}
\caption{\small Prompting details of the \method\ method. The green texts need to be filled with the input questions or the arguments and outputs of other agents with the JSON format.}
\label{fig:ma_prompts}
\end{figure*}

In \dataset, we only provide the test set for the LLMs evaluation without training or fine-tuning. So traditional language models such as RNN or LSTM~\cite{rumelhart1986learning-rnn,hochreiter1997long-lstm,wang2023tensor,DBLP:conf/aaai/PanXWYWBX19} are not evaluated. For GPT3.5, we use OpenAI GPT-3.5-Turbo API. \textcolor{black}{Specifically, we use GPT-3.5-Turbo-0613. Though previous studies show that the output of the API may change over time\citep{tu2023chatlog,chen2023chatgpts}, we are able to reproduce the results.} While for LLaMA2, we use 4-bit quantization to save the memory. For the generation output of the LLMs, we first parse the output heuristically. We try to find the string after the phrase "answer is", where we try to match the choices. If this pre-parse fails, we use Levenshtein distance to get the final choice. The detailed prompts are shown in Figure~\ref{fig:ma_prompts}.

\subsection{Detailed Experimental Analysis}
\begin{figure}[hbt]
\centering
    \includegraphics[width=0.9\linewidth]{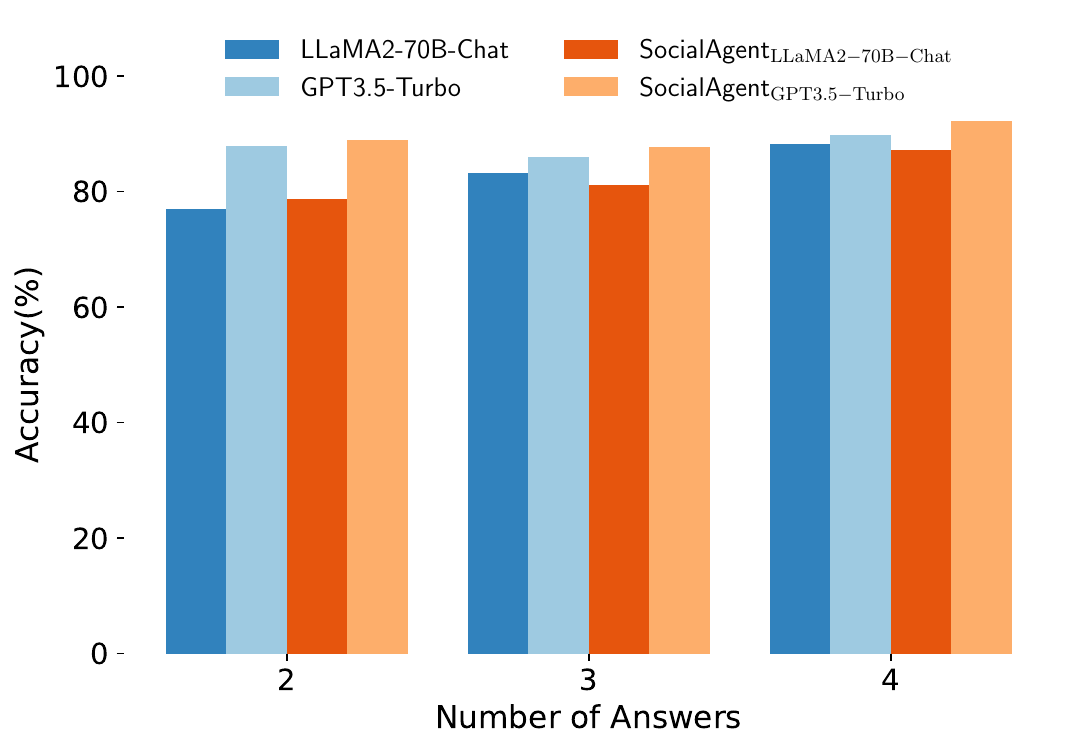}
    \caption{\small Results on questions with different numbers of answers.}
    \label{fig:number_of_answers}
\end{figure}

\paragraph{Number of Answers}
We also analyze how model performance changes with the number of answers. We show the results in Figure~\ref{fig:number_of_answers}. Surprisingly, the accuracy increases with the number of answers, which is contrary to the expectation that more choices lead to harder problems.

\begin{figure}[!tbh]
    \centering
    \centering
    \includegraphics[width=\linewidth]{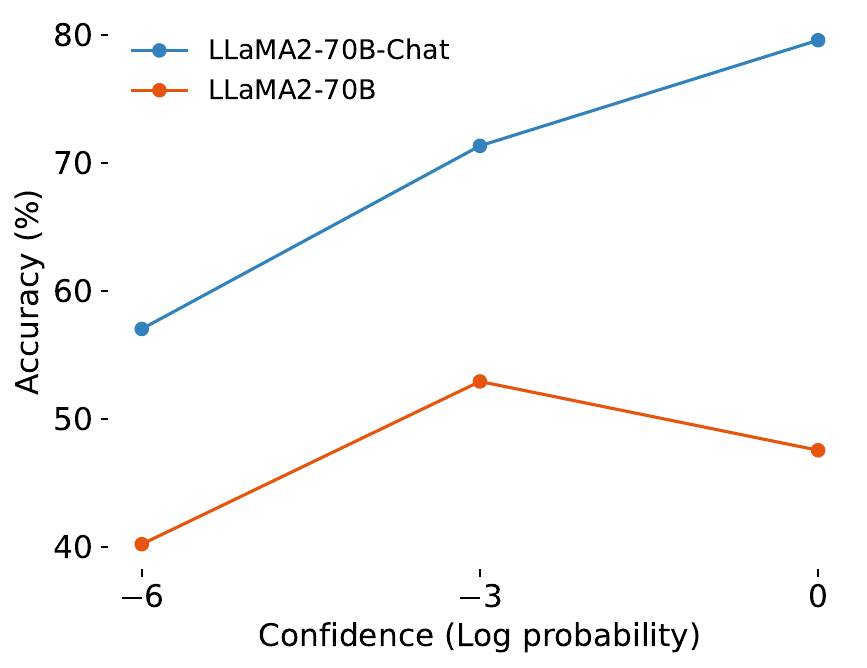}
    \caption{\small LLaMA2 calibration results. The $x$-axis denotes the sum of the log probability of the predicted answers.}
    \label{fig:calibration}
\end{figure}

\paragraph{Calibration}
We show the relationship between the confidence of LLaMA2-70B models and the corresponding accuracy in Figure~\ref{fig:calibration}. A reliable model should be calibrated, which means the output confidence should match the accuracy~\cite{calibration}. We use the sum of the log probability of the predicted answers as the confidence and show the relationship between the confidence and the accuracy. We find that the zero-shot LLaMA2-70B-Chat is well-calibrated. However, the pretrained LLaMA2-70B model without instruction tuning is not well-calibrated, which shows the importance of alignment fine-tuning.

\paragraph{More Details on Exam Score}
We show the Exam Score for each skill in Table~\ref{ela_detailed_smart_score1} to ~\ref{ela_detailed_smart_score4}. As demonstrated in the table, our \method\ method can achieve 100 exam scores in a significant amount of skills, even for some skills that the other three methods get lower performance. 

\begin{figure}[hbt]
\centering
    \includegraphics[width=0.9\linewidth]{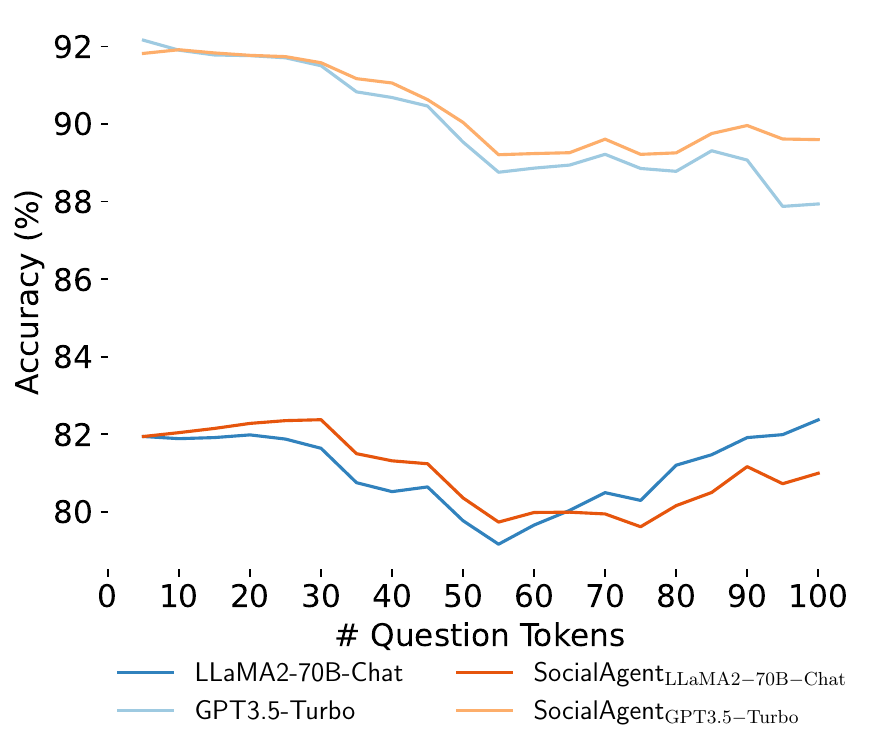}
    \caption{\small Results on questions with different lengths.}
    \label{fig:ques_len}
\end{figure}

\paragraph{Question Lengths}
We show how the question length affects model accuracy in Figure~\ref{fig:ques_len}. Overall, we can find that the longer the question length, the harder the question, and the worse the performance. Moreover, it can be discovered that the curves for \method\ are smoother than zero-shot settings, which means for a better model, the difficulty of the question does not interfere with it more.

\begin{figure}[hbt]
\centering
    \includegraphics[width=0.9\linewidth]{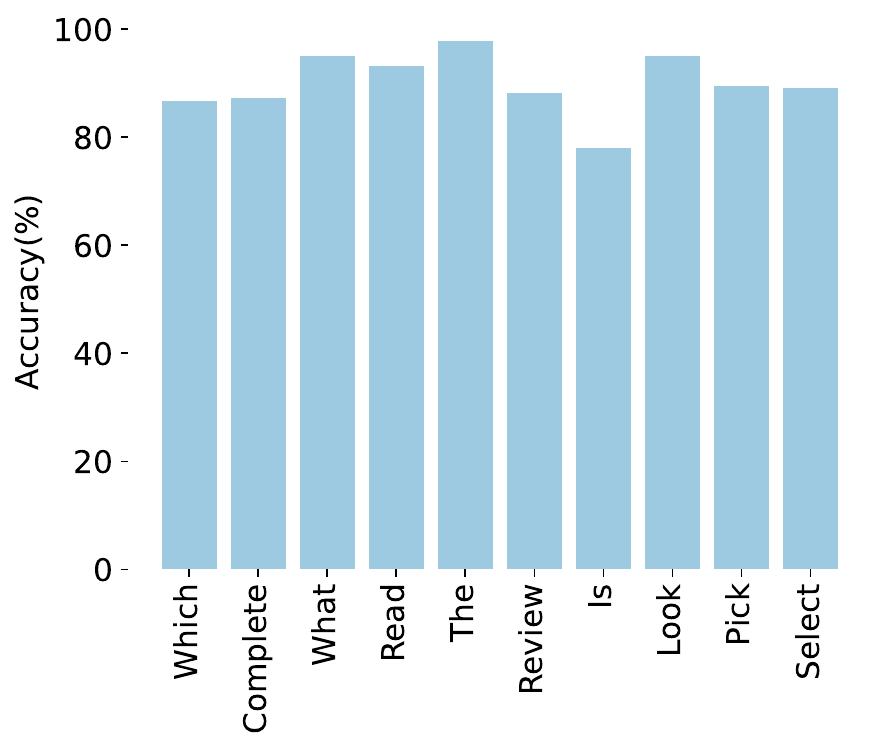}
    \caption{\small Results on different question types.}
    \label{fig:ques_type}
\end{figure}

\paragraph{Question Type}
We use the first word in the question to mark the types of problems, and list the accuracy on the top-10 number of question types in Figure~\ref{fig:ques_type}. Questions starting with ``Is'' have relatively low accuracy, which means they are harder to answer. 

\begin{figure}[hbt]
\centering
    \includegraphics[width=0.9\linewidth]{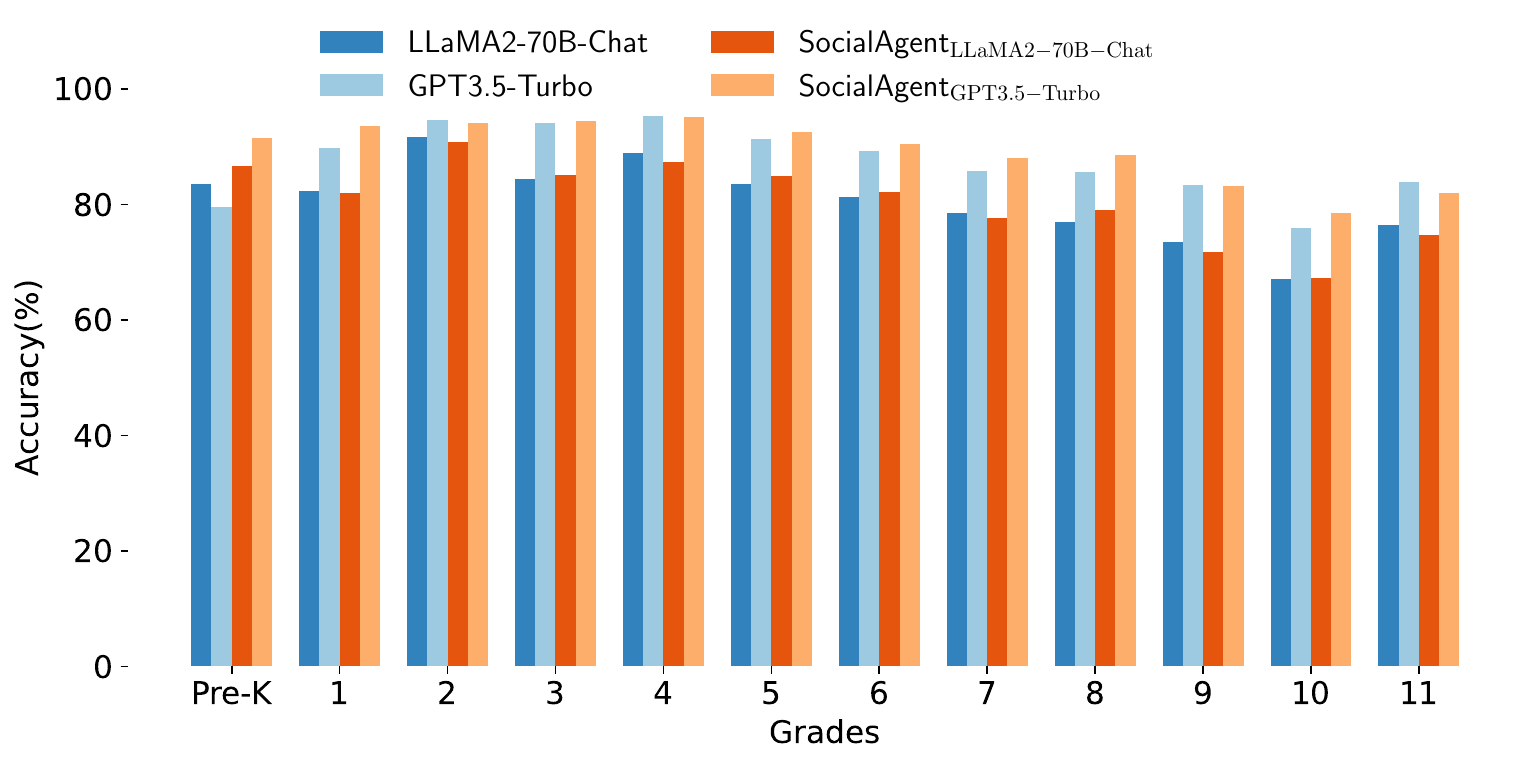}
    \caption{Average grade-level accuracies.}
    \label{fig:grade_accuracy}
\end{figure}

\paragraph{Grades} 
We show the accuracies of the models along each grade in Figure~\ref{fig:grade_accuracy}. We find that the higher the grade, the lower accuracy of the models. Moreover, the \method\ method can help the agents perform better in each grade.

\paragraph{Hard Questions}
\textcolor{black}{Models in general obtain the lowest score in tenth grade questions. This means that the higher the grade, the harder the questions are. Tenth grade questions are hardest for the models. Besides, we also find skills such as ``understand-overall-supply-and-demand'' are hard for LLMs with an average of 71.0\% accuracy (much lower than the average accuracy).} 

\paragraph{Data Contamination}
\textcolor{black}{The source datasets (e.g., IXL) require registration to access their data and are designed for education purposes. So it is very unlikely that the data is part of the training data of the LLMs. In addition, we carefully checked the “Data Contamination” section in the technical reports of GPT-3 and LLaMA-2, and it seems the contamination is currently not a major issue of the performance.}

\begin{figure}[ht]
\centering
    \includegraphics[width=0.9\linewidth]{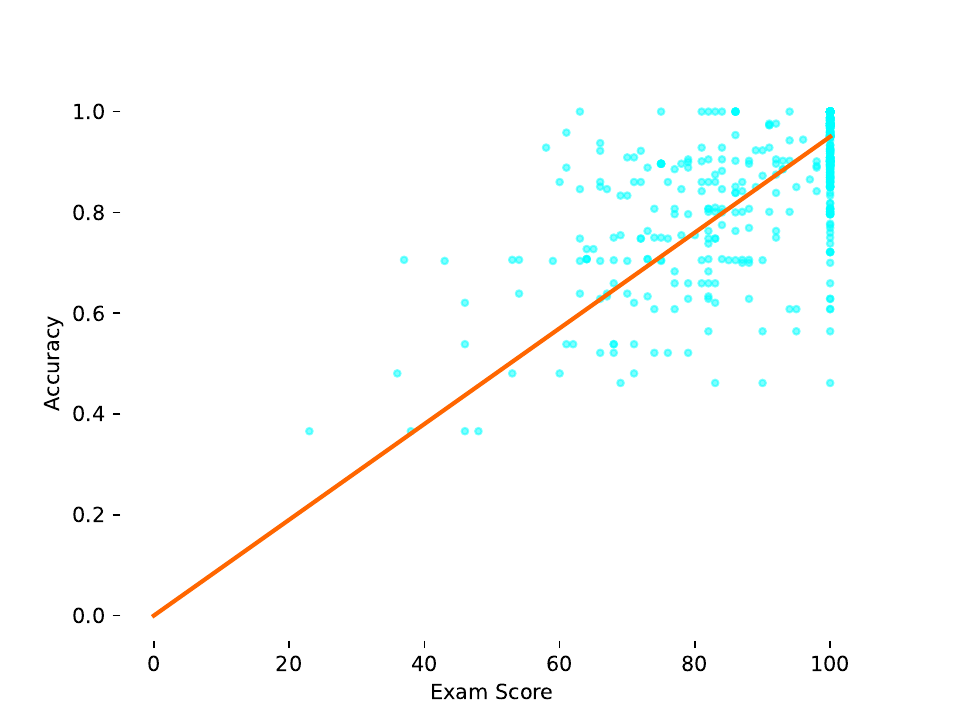}
    \caption{The correlation graph of exam scores with model accuracies.}
    \label{fig:smart_acc}
\end{figure}

\paragraph{Correlation Analysis}
We evaluate exam scores' correlation with the model accuracies in Figure~\ref{fig:smart_acc}. A positive correlation can be found, and exam scores can capture the accuracy as an important factor.

\begin{table}[ht]
    \centering
\resizebox{0.95\linewidth}{!}{
    \begin{tabular}{|c|c|}
    \hline
    \method\ Component & Inputs and Outputs \\ \hline
    Problem &
    \scalebox{0.75}{\begin{minipage}{300pt}
        \quad \\
        Which is an example of a law? \\
        (a) You should get a new toothbrush every six months. \\
        (b) You may not drive above the speed limit. \\
        (c) You must clean your room before dinner. \\
        (d) You should eat more fruits and vegetables. \\
    \end{minipage}}\\ \hline
    Retrieval Agent &
    \scalebox{0.75}{\begin{minipage}{300pt}
        \quad \\
\textbf{Input:} law \\
\textbf{Output:} Page: Law\\
Summary: Law is a set of rules that are created and are enforceable by social or governmental institutions to regulate behavior, with its precise definition a matter of longstanding debate. It has been variously described as a science and as the art of justice. State-enforced laws can be made by a group legislature or by a single legislator, resulting in statutes; by the executive through decrees and regulations; or established by judges through precedent, usually in common law jurisdictions. Private individuals may create legally binding contracts, including arbitration agreements that adopt alternative ways of resolving disputes to standard court litigation. The creation of laws themselves may be influenced by a constitution, written or tacit, and the rights encoded therein. The law shapes politics, economics, history and society in various ways and also serves as a mediator of relations between people.\\
Legal systems vary between jurisdictions, with their differences analysed in comparative law. \\
\textbf{Useful:} true \\
    \end{minipage}}\\ \hline
    Programming Agent &
    \scalebox{0.75}{\begin{minipage}{300pt}
        \quad \\
\textbf{Input:} ``''+``'' \\
\textbf{Output:} NA \\
\textbf{Useful:} false \\
    \end{minipage}}\\ \hline
    Reasoning Agent &
     \scalebox{0.75}{\begin{minipage}{300pt}
        \quad \\
\textbf{Output:} To determine which of the given options is an example of a law, we need to understand the definition of a law. According to the information provided by the Wikipedia search, a law is a set of rules created and enforceable by social or governmental institutions to regulate behavior.\\
Let's analyze each option:\\
(a) You should get a new toothbrush every six months. This statement is a recommendation or advice, but it does not represent a law. It does not involve any enforcement by social or governmental institutions.\\
(b) You may not drive above the speed limit. This statement represents a law. It sets a rule that prohibits driving above the speed limit, and it is enforceable by traffic laws and law enforcement agencies.\\
(c) You must clean your room before dinner. This statement is a rule or expectation, but it does not represent a law. It is more likely a household rule or expectation rather than a law enforced by social or governmental institutions.\\
(d) You should eat more fruits and vegetables. Similar to option (a), this statement is a recommendation or advice, but it does not represent a law. It does not involve any enforcement by social or governmental institutions.\\
Based on the analysis, the example of a law among the given options is (b) You may not drive above the speed limit.\\
\textbf{Useful:} true \\
    \end{minipage}}\\ \hline
    Answer &
    \scalebox{0.75}{\begin{minipage}{300pt}
        \quad \\
        \textbf{Predicted String:} (a) surplus\\
    \end{minipage}}\\ \hline
    \end{tabular}
}
    \caption{Full example of Figure~\ref{fig:ma_prompts}.}
    \label{tab:main_full}
\end{table}

\begin{figure*}[!t]
    \centering
    \begin{minipage}[t]{0.42\linewidth}

\centering
\resizebox{0.95\linewidth}{!}{
    \begin{tabular}{|c|c|}
    \hline
    \method\ Component & Inputs and Outputs \\ \hline
    Problem &
    \scalebox{0.75}{\begin{minipage}{300pt}
        \quad \\
        What month comes right before February? \\
        (a) August \\
        (b) January \\

    \end{minipage}}\\ \hline
    Retrieval Agent &
    \scalebox{0.75}{\begin{minipage}{300pt}
        \quad \\
\textbf{Input:} month comes right before February \\
\textbf{Output:} Page: Before Sunrise \\
Summary: Before Sunrise is a 1995 romantic drama film directed by Richard Linklater and co-written by Linklater and Kim Krizan… Before Sunrise also explores time, death, and self-discovery. \\
\textbf{Useful:} false \\
    \end{minipage}}\\ \hline
    Programming Agent &
    \scalebox{0.75}{\begin{minipage}{300pt}
        \quad \\
\textbf{Input:} ``''+``'' \\
\textbf{Output:} NA \\
\textbf{Useful:} false \\
    \end{minipage}}\\ \hline
    Reasoning Agent &
     \scalebox{0.75}{\begin{minipage}{300pt}
        \quad \\
\textbf{Output:} To determine the month that comes right before February, we can use our knowledge of the calendar. \\
Step 1: Recall the order of the months in a calendar year: January, February, March, April, May, June, July, August, September, October, November, December. \\
Step 2: Identify the month that comes immediately before February in the calendar year. \\
Based on the given choices, the correct answer is (b) January. \\
We can disregard the information from the other agents as it does not provide any relevant information about the order of the months. \\
\textbf{Useful:} true \\
    \end{minipage}}\\ \hline
    Answer &
    \scalebox{0.75}{\begin{minipage}{300pt}
        \quad \\
        \textbf{Predicted String:} (b) January\\
    \end{minipage}}\\ \hline
    \end{tabular}
}

    \end{minipage}
    \hfill
    \begin{minipage}[t]{0.42\linewidth}
        \centering
\centering
\resizebox{0.95\linewidth}{!}{
    \begin{tabular}{|c|c|}
    \hline
    \method\ Component & Inputs and Outputs \\ \hline
    Problem &
    \scalebox{0.75}{\begin{minipage}{300pt}
        \quad \\
        In most state legislatures, the upper house is called the Senate and the lower house is called the House of Representatives. Is the House of Representatives bigger or smaller than the Senate? \\
        (a) The House of Representatives is smaller. \\
        (b) The House of Representatives is bigger. \\

    \end{minipage}}\\ \hline
    Retrieval Agent &
    \scalebox{0.75}{\begin{minipage}{300pt}
        \quad \\
\textbf{Input:} size of House of Representatives compared to Senate \\
\textbf{Output:} Page: House of Representatives of the Philippines \\
Summary: The House of Representatives of the Philippines (Filipino: Kapulungan ng mga Kinatawan ng Pilipinas… The 19th Congress has 253 congressional districts. Party-list representatives are elected...
\textbf{Useful:} true \\
    \end{minipage}}\\ \hline
    Programming Agent &
    \scalebox{0.75}{\begin{minipage}{300pt}
        \quad \\
\textbf{Input:} ``''+``'' \\
\textbf{Output:} NA \\
\textbf{Useful:} false \\
    \end{minipage}}\\ \hline
    Reasoning Agent &
     \scalebox{0.75}{\begin{minipage}{300pt}
        \quad \\
\textbf{Output:} To determine whether the House of Representatives is bigger or smaller than the Senate, we can use the information provided and analyze it step-by-step. \\
1. The question states that in most state legislatures, the upper house is called the Senate and the lower house is called the House of Representatives. \\
2. We can refer to the information provided by the other agent, which states that the House of Representatives is the lower house of Congress…  \\
Therefore, the answer is: (a) The House of Representatives is smaller. \\
\textbf{Useful:} true \\
    \end{minipage}}\\ \hline
    Answer &
    \scalebox{0.75}{\begin{minipage}{300pt}
        \quad \\
        \textbf{Predicted String:} (a) The House of Representatives is smaller. \\
    \end{minipage}}\\ \hline
    \end{tabular}
}
    \end{minipage}
    \caption{Complete results of Case Study (Figure~\ref{fig:case_study}).}
    \label{fig:case_study_full}
\end{figure*}

\section{Prompting Details of \method\ Method}
\label{app:multi_agent_prompting}

We show the prompts of our \method\ method in Figure~\ref{fig:ma_prompts}. And we show the complete \method\ results of Figure~\ref{fig:multi_agents} and in Table~\ref{tab:main_full}. \textcolor{black}{In the prompt design, we aim to help models avoid redundant information such as ``Sure, I'm glad to help you''. We therefore do not require models to be polite but be concise regarding their responses.}
We also show the complete results of our case study~\ref{fig:case_study} in Table~\ref{fig:case_study_full}.

\section{Summary of Skills}
\label{apped:summary_skills}
We list all the skills in Table~\ref{tab:skill_summary_full_social} to~\ref{tab:skill_summary_full_ela2}, and show an example for each skill in Table~\ref{tab:example_by_skill_1} to~\ref{tab:example_by_skill_26}.

\newpage
\begin{table*}
    \centering
\scalebox{0.43}{
\begin{tabular}{>{\centering\arraybackslash}p{14cm}|>{\centering\arraybackslash}p{3.5cm}|>{\centering\arraybackslash}p{3.5cm}|>{\centering\arraybackslash}p{5.5cm}|>{\centering\arraybackslash}p{3.5cm}|>{\centering\arraybackslash}p{5.5cm}}
    \toprule
    Grade/Skill & Random & LLaMA2-70B-Chat & \method$_{\rm LLaMA2-70B-Chat}$ & GPT3.5-Turbo & \method$_{\rm GPT3.5-Turbo}$ \\
    \hline
pre-k/choose-the-uppercase-letter-that-matches-c-k-o-p-s-u-v-w-x-z & 70 & 100 & 100 & 100 & 100 \\
pre-k/choose-the-lowercase-letter-that-matches-c-k-o-p-s-u-v-w-x-z & 70 & 100 & 100 & 100 & 100 \\
pre-k/choose-the-lowercase-letter-that-matches-f-i-j-l-m-t-y & 50 & 100 & 100 & 100 & 100 \\
pre-k/choose-the-lowercase-letter-that-matches-a-b-d-e-g-h-n-q-r & 76 & 100 & 100 & 100 & 100 \\
pre-k/choose-the-uppercase-letter-that-matches-a-b-d-e-g-h-n-q-r & 76 & 100 & 100 & 100 & 100 \\
pre-k/choose-the-uppercase-letter-that-matches-f-i-j-l-m-t-y & 80 & 100 & 100 & 100 & 100 \\
kindergarten/choose-the-lowercase-letter-that-matches-c-k-o-p-s-u-v-w-x-z & 41 & 100 & 100 & 100 & 100 \\
\hline
    \end{tabular}
    }
    \caption{Exam scores for Language Arts skill (part 1).}
    \label{ela_detailed_smart_score1}
\end{table*}

\begin{table*}
    \centering
\scalebox{0.43}{
    \begin{tabular}{>{\centering\arraybackslash}p{14cm}|>{\centering\arraybackslash}p{3.5cm}|>{\centering\arraybackslash}p{3.5cm}|>{\centering\arraybackslash}p{5.5cm}|>{\centering\arraybackslash}p{3.5cm}|>{\centering\arraybackslash}p{5.5cm}}
    \toprule
    Grade/Skill & Random & LLaMA2-70B-Chat & \method$_{\rm LLaMA2-70B-Chat}$ & GPT3.5-Turbo & \method$_{\rm GPT3.5-Turbo}$ \\
    \hline
grade-4/describe-the-difference-between-related-words & 52 & 89 & 88 & 100 & 100 \\
grade-4/to-have-use-the-correct-form & 63 & 100 & 100 & 100 & 100 \\
grade-4/use-subordinating-conjunctions & 49 & 100 & 64 & 100 & 100 \\
grade-4/use-the-correct-homophone & 39 & 100 & 100 & 100 & 100 \\
grade-4/is-it-a-complete-sentence-or-a-fragment & 67 & 70 & 97 & 100 & 100 \\
grade-4/choose-the-best-concluding-sentence & 46 & 100 & 100 & 100 & 100 \\
grade-4/identify-the-narrative-point-of-view & 52 & 88 & 76 & 82 & 100 \\
\hline
    \end{tabular}
    }
    \caption{Exam scores for Language Arts skill (part 2).}
    \label{ela_detailed_smart_score2}
\end{table*}

\begin{table*}
    \centering
\scalebox{0.43}{
    \begin{tabular}{>{\centering\arraybackslash}p{14cm}|>{\centering\arraybackslash}p{3.5cm}|>{\centering\arraybackslash}p{3.5cm}|>{\centering\arraybackslash}p{5.5cm}|>{\centering\arraybackslash}p{3.5cm}|>{\centering\arraybackslash}p{5.5cm}}
    \toprule
    Grade/Skill & Random & LLaMA2-70B-Chat & \method$_{\rm LLaMA2-70B-Chat}$ & GPT3.5-Turbo & \method$_{\rm GPT3.5-Turbo}$ \\
    \hline
grade-6/is-the-pronoun-reflexive-or-intensive & 47 & 56 & 46 & 63 & 74 \\
grade-6/interpret-figures-of-speech & 61 & 100 & 100 & 100 & 100 \\
grade-6/classify-figures-of-speech & 76 & 100 & 100 & 81 & 80 \\
grade-6/is-the-sentence-declarative-interrogative-imperative-or-exclamatory & 31 & 73 & 79 & 100 & 100 \\
grade-6/vocabulary-review-trace-an-argument & 40 & 100 & 100 & 100 & 100 \\
grade-6/identify-the-authors-purpose & 39 & 100 & 100 & 100 & 100 \\
grade-6/is-it-a-direct-object-or-an-indirect-object & 44 & 49 & 71 & 70 & 100 \\
\hline
    \end{tabular}
    }
    \caption{Exam scores for Language Arts skill (part 3).}
    \label{ela_detailed_smart_score3}
\end{table*}

\begin{table*}
    \centering
\scalebox{0.43}{
    \begin{tabular}{>{\centering\arraybackslash}p{14cm}|>{\centering\arraybackslash}p{3.5cm}|>{\centering\arraybackslash}p{3.5cm}|>{\centering\arraybackslash}p{5.5cm}|>{\centering\arraybackslash}p{3.5cm}|>{\centering\arraybackslash}p{5.5cm}}
    \toprule
    Grade/Skill & Random & LLaMA2-70B-Chat & \method$_{\rm LLaMA2-70B-Chat}$ & GPT3.5-Turbo & \method$_{\rm GPT3.5-Turbo}$ \\
    \hline
grade-8/is-the-pronoun-reflexive-or-intensive & 43 & 64 & 70 & 57 & 92 \\
grade-8/interpret-figures-of-speech & 59 & 100 & 100 & 100 & 100 \\
grade-8/identify-the-verb-mood & 32 & 76 & 56 & 100 & 100 \\
grade-8/classify-figures-of-speech & 64 & 100 & 84 & 100 & 78 \\
grade-8/is-the-sentence-declarative-interrogative-imperative-or-exclamatory & 36 & 83 & 94 & 100 & 100 \\
grade-8/vocabulary-review-trace-an-argument & 37 & 84 & 100 & 100 & 100 \\
grade-8/is-it-a-direct-object-or-an-indirect-object & 60 & 63 & 79 & 75 & 100 \\
\hline
    \end{tabular}
    }
    \caption{Exam scores for Language Arts skill (part 4).}
    \label{ela_detailed_smart_score4}
\end{table*}

\begin{table*}
    \centering
\scalebox{0.43}{
    \begin{tabular}{>{\centering\arraybackslash}p{17cm}|>{\centering\arraybackslash}p{2.5cm}|>{\centering\arraybackslash}p{3.4cm}|>{\centering\arraybackslash}p{5.4cm}|>{\centering\arraybackslash}p{3.1cm}|>{\centering\arraybackslash}p{5.1cm}}
    \toprule
    Grade/Skill & Random & LLaMA2-70B-Chat & \method$_{\rm LLaMA2-70B-Chat}$ & GPT3.5-Turbo & \method$_{\rm GPT3.5-Turbo}$ \\
    \hline
grade-10/use-thesaurus-entries & 68 & 100 & 100 & 100 & 100 \\
grade-10/determine-the-main-idea-of-a-passage & 53 & 100 & 11 & 100 & 100 \\
grade-10/determine-the-meanings-of-greek-and-latin-roots & 28 & 100 & 90 & 100 & 100 \\
grade-10/use-parallel-structure & 76 & 75 & 66 & 70 & 100 \\
grade-10/use-dictionary-entries-to-determine-correct-usage & 55 & 18 & 85 & 49 & 93 \\
grade-10/which-text-is-most-formal & 40 & 100 & 100 & 100 & 100 \\
grade-10/choose-the-topic-sentence-that-best-captures-the-main-idea & 37 & 39 & 93 & 100 & 100 \\
\hline
    \end{tabular}
    }
    \caption{Exam scores for Language Arts skill (part 5).}
    \label{ela_detailed_smart_score5}
\end{table*}
\begin{table*}
    \centering
\scalebox{0.43}{
    \begin{tabular}{>{\centering\arraybackslash}p{14cm}|>{\centering\arraybackslash}p{3.5cm}|>{\centering\arraybackslash}p{3.5cm}|>{\centering\arraybackslash}p{5.5cm}|>{\centering\arraybackslash}p{3.5cm}|>{\centering\arraybackslash}p{5.5cm}}
    \toprule
    Grade/Skill & Random & LLaMA2-70B-Chat & \method$_{\rm LLaMA2-70B-Chat}$ & GPT3.5-Turbo & \method$_{\rm GPT3.5-Turbo}$ \\
    \hline
grade-2/compare-urban-suburban-and-rural-areas & 29 & 100 & 100 & 100 & 100 \\
grade-2/rosa-parks & 0 & 67 & 60 & 73 & 73 \\
grade-3/thurgood-marshall & 0 & 96 & 96 & 96 & 96 \\
grade-3/rosa-parks & 6 & 61 & 56 & 61 & 61 \\
grade-3/the-bill-of-rights & 56 & 100 & 100 & 100 & 100 \\
grade-4/the-war-of-1812-events-and-effects & 3 & 66 & 69 & 72 & 66 \\
grade-4/understand-overall-supply-and-demand & 33 & 100 & 100 & 100 & 100 \\
grade-5/the-war-of-1812-causes & 0 & 63 & 58 & 63 & 63 \\
grade-5/the-war-of-1812-events-and-effects & 0 & 66 & 66 & 72 & 72 \\
grade-5/the-bill-of-rights & 26 & 100 & 100 & 100 & 100 \\
grade-5/understand-overall-supply-and-demand & 14 & 100 & 100 & 100 & 100 \\
grade-5/identify-shortage-and-surplus & 29 & 40 & 30 & 50 & 100 \\
grade-6/understand-overall-supply-and-demand & 49 & 100 & 100 & 100 & 100 \\
grade-7/the-bill-of-rights & 31 & 100 & 100 & 100 & 100 \\
grade-8/the-bill-of-rights & 31 & 95 & 100 & 100 & 100 \\
grade-8/understand-overall-supply-and-demand & 40 & 100 & 100 & 100 & 100 \\
grade-8/identify-primary-and-secondary-sources & 37 & 75 & 75 & 75 & 75 \\
grade-8/identify-facts-and-opinions & 59 & 100 & 100 & 100 & 100 \\
    \hline
    \end{tabular}
    }
    \caption{Exam scores for Social Studies skill.}
    \label{social_detailed_smart_score}
\end{table*}

\begin{table*}
    \centering
\scalebox{0.4}{
    \begin{tabular}{c|l}
    \toprule
    Grade & \multicolumn{1}{c}{Skills of Social Studies} \\
    \midrule
    
	kindergarten & \makecell[l]{\\days-of-the-week, pledge-of-allegiance, months-of-the-year\\\ } \\
	\hline
	grade-1 & \makecell[l]{\\months-of-the-year, days-of-the-week\\\ } \\
	\hline
	grade-2 & \makecell[l]{\\hanukkah, kwanzaa, compare-urban-suburban-and-rural-areas, thanksgiving, sitting-bull, susan-b-anthony, harriet-tubman, neil-armstrong, thurgood-marshall, federal-government, ramadan, easter, cesar-ch\\avez, jackie-robinson, rosh-hashanah, amelia-earhart, christmas, dia-de-los-muertos, paul-revere, the-statue-of-liberty, george-washington-carver, davy-crockett, thomas-jefferson, state-government, lu\\nar-new-year, bill-gates, the-white-house, abraham-lincoln, john-deere, holi, rosa-parks, frederick-douglass, benjamin-franklin, local-government, purpose-of-government, theodore-roosevelt\\\ } \\
	\hline
	grade-3 & \makecell[l]{\\easter, theodore-roosevelt, state-government, holi, cesar-chavez, the-bill-of-rights, bill-gates, ramadan, the-constitution, frederick-douglass, thomas-edison, jackie-robinson, kwanzaa, christmas, sus\\an-b-anthony, harriet-tubman, paul-revere, davy-crockett, thurgood-marshall, george-washington-carver, sitting-bull, neil-armstrong, thomas-jefferson, hanukkah, dia-de-los-muertos, lunar-new-year, ros\\a-parks, thanksgiving, presidential-elections\\\ } \\
	\hline
	grade-4 & \makecell[l]{\\jamestown-growth-of-a-colony, understand-overall-supply-and-demand, the-civil-war-war-tactics-and-the-home-front, the-war-of-1812-events-and-effects, the-constitution, the-american-revolution-the-rebe\\llion-begins, world-war-i-the-road-to-peace, costs-and-benefits, new-england-colonies-religion-and-government, the-war-of-1812-causes\\\ } \\
\hline

    \end{tabular}
    }
    \caption{Full social studies skill summary.}
    \label{tab:skill_summary_full_social}
\end{table*}

\begin{table*}
    \centering
\scalebox{0.4}{
    \begin{tabular}{c|l}
    \toprule
    Grade & \multicolumn{1}{c}{Skills of Language Arts} \\
    \midrule
    pre-k & \makecell[l]{\\choose-the-uppercase-letter-that-matches-c-k-o-p-s-u-v-w-x-z, choose-the-lowercase-letter-that-matches-c-k-o-p-s-u-v-w-x-z, choose-the-lowercase-letter-that-matches-f-i-j-l-m-t-y, choose-the-uppercase\\-letter-that-matches-f-i-j-l-m-t-y, choose-the-uppercase-letter-that-matches-a-b-d-e-g-h-n-q-r, choose-the-lowercase-letter-that-matches-a-b-d-e-g-h-n-q-r\\\ } \\
	\hline
	kindergarten & \makecell[l]{\\who-what-when-where-or-why, complete-the-sentence-with-the-correct-sight-word, find-the-vowel-in-the-word, choose-the-uppercase-letter-that-matches-f-i-j-l-m-t-y, choose-the-uppercase-letter-that-matc\\hes-review, choose-the-lowercase-letter-that-matches-a-b-d-e-g-h-n-q-r, choose-the-lowercase-letter-that-matches-review, capitalize-the-first-letter-of-a-sentence, complete-the-sentence-with-the-corre\\ct-short-vowel-word, choose-the-lowercase-letter-that-matches-c-k-o-p-s-u-v-w-x-z, read-questions-with-short-vowel-words, find-the-complete-sentence, choose-the-uppercase-letter-that-matches-c-k-o-p-s\\-u-v-w-x-z, find-the-action-verb, is-it-a-telling-sentence-or-an-asking-sentence, complete-the-rhyme, which-word-does-not-rhyme, choose-the-lowercase-letter-that-matches-f-i-j-l-m-t-y, what-am-i, choo\\se-the-uppercase-letter-that-matches-a-b-d-e-g-h-n-q-r\\\ } \\
\hline

    \end{tabular}
    }
    \caption{Full language arts skill summary (part 1).}
    \label{tab:skill_summary_full_ela1}
\end{table*}

\begin{table*}
    \centering
\scalebox{0.4}{
    \begin{tabular}{c|l}
    \toprule
    Grade & \multicolumn{1}{c}{Skills of Language Arts} \\
\hline

    \end{tabular}
    }
    \caption{Full language arts skill summary (part 2).}
    \label{tab:skill_summary_full_ela2}
\end{table*}
\begin{table*}
    \centering
\scalebox{0.6}{
    \begin{tabular}{|c|c|}
    \hline
    \scalebox{0.8}{\begin{minipage}{400pt}
        Subject: Social Studies \\
        Skill: the-american-revolution-new-british-taxes \\
        Description: In the 1760s, smuggling was an important part of the colonial economy. What does smuggling mean? \\
        Choices: [(a) stealing goods from foreign merchants, (b) trading one good for another without the use of money, \textcolor{red}{(c) shipping goods secretly or illegally}, (d) producing metal from minerals and rocks]\\
        Answer index: 2 \\
    \end{minipage}}
&
    \scalebox{0.8}{\begin{minipage}{400pt}
        Subject: Social Studies \\
        Skill: world-war-i-the-road-to-war \\
        Description: What was World War I originally called? \\
        Choices: [(a) the First War, \textcolor{red}{(b) the Great War}, (c) the European Revolution, (d) European War I]\\
        Answer index: 1 \\
    \end{minipage}}\\
\hline
    \scalebox{0.8}{\begin{minipage}{400pt}
        Subject: Social Studies \\
        Skill: the-american-flag \\
        Description: What does each star on the American flag stand for? \\
        Choices: [(a) a president, \textcolor{red}{(b) a state}, (c) a war, (d) a city]\\
        Answer index: 1 \\
    \end{minipage}}
&
    \scalebox{0.8}{\begin{minipage}{400pt}
        Subject: Social Studies \\
        Skill: identify-facts-and-opinions \\
        Description: Which sentence states a fact? \\
        Choices: [\textcolor{red}{(a) The Grand Canyon was carved by the Colorado River.}, (b) The Grand Canyon is the most breathtaking place the United States.]\\
        Answer index: 0 \\
    \end{minipage}}\\
\hline
    \end{tabular}
}
    \caption{Question examples for each skill (part 1).}
    \label{tab:example_by_skill_1}
\end{table*}
\begin{table*}
    \centering
\scalebox{0.6}{
    \begin{tabular}{|c|c|}
    \hline
    \scalebox{0.8}{\begin{minipage}{400pt}
        Subject: Social Studies \\
        Skill: identify-primary-and-secondary-sources \\
        Description: Imagine that you are a historian who wants to learn more about the fall of the Roman Empire in the 400s CE. You find the source described below. Read the description of the source. Then complete the sentence below. \\The historian Edward Gibbon wrote the book The History of the Decline and Fall of the Roman Empire in 1776 CE. \\Complete the sentence. \\In this case, the book is a \_\_\_ source because it \_\_\_. \\
        Choices: [(a) primary . . . is a firsthand account of the fall of the Roman Empire, \textcolor{red}{(b) secondary . . . was written by a historian after the fall of the Roman Empire}]\\
        Answer index: 1 \\
    \end{minipage}}
&
    \scalebox{0.8}{\begin{minipage}{400pt}
        Subject: Social Studies \\
        Skill: months-of-the-year \\
        Description: What month comes right after February? \\
        Choices: [\textcolor{red}{(a) March}, (b) December]\\
        Answer index: 0 \\
    \end{minipage}}\\
\hline
    \scalebox{0.8}{\begin{minipage}{400pt}
        Subject: Social Studies \\
        Skill: days-of-the-week \\
        Description: What day comes right after Tuesday? \\
        Choices: [(a) Saturday, (b) Friday, \textcolor{red}{(c) Wednesday}]\\
        Answer index: 2 \\
    \end{minipage}}
&
    \scalebox{0.8}{\begin{minipage}{400pt}
        Subject: Social Studies \\
        Skill: theodore-roosevelt \\
        Description: Complete the sentence. \\Theodore Roosevelt is best known as \_\_\_ from the early 1900s.  \\
        Choices: [\textcolor{red}{(a) a president}, (b) an artist, (c) a pilot, (d) an inventor]\\
        Answer index: 0 \\
    \end{minipage}}\\
\hline
    \end{tabular}
}
    \caption{Question examples for each skill (part 2).}
    \label{tab:example_by_skill_2}
\end{table*}
\begin{table*}
    \centering
\scalebox{0.6}{
    \begin{tabular}{|c|c|}
    \hline
    \scalebox{0.8}{\begin{minipage}{400pt}
        Subject: Social Studies \\
        Skill: the-great-depression-part-i \\
        Description: Who was the president of the United States at the start of the Great Depression? \\
        Choices: [(a) William McKinley, \textcolor{red}{(b) Herbert Hoover}, (c) Harry Truman, (d) Gerald Ford]\\
        Answer index: 1 \\
    \end{minipage}}
&
    \scalebox{0.8}{\begin{minipage}{400pt}
        Subject: Social Studies \\
        Skill: the-american-revolution-struggle-for-independence \\
        Description: At the Battle of Bunker Hill, one of the colonial officers gave the soldiers a famous order. Read the order. Then answer the question below. \\Don't fire until you can see the whites of their eyes! \\What was one reason the officer gave this order? \\
        Choices: [\textcolor{red}{(a) He did not want the soldiers to waste any gunpowder or ammunition.}, (b) He wanted to make sure that the British soldiers were really going to attack., (c) He wanted to make sure that the soldiers were actually British., (d) He did not want the British soldiers to be surprised by the gunfire.]\\
        Answer index: 0 \\
    \end{minipage}}\\
\hline
    \end{tabular}
}
    \caption{Question examples for each skill (part 3).}
    \label{tab:example_by_skill_3}
\end{table*}
\begin{table*}
    \centering
\scalebox{0.6}{
    \begin{tabular}{|c|c|}
    \hline
    \scalebox{0.8}{\begin{minipage}{400pt}
        Subject: Social Studies \\
        Skill: antebellum-period-slavery-and-politics-part-ii \\
        Description: While in prison in December 1859, John Brown handed a note to his guard. Use the text from that note to answer the question below. \\I, John Brown, am now quite certain that the crimes of this guilty land will never be purged away; but with blood. \\purged:removed \\What did Brown believe would be needed to end slavery? \\
        Choices: [(a) economic boycott, (b) petitions, (c) non-violent protest, \textcolor{red}{(d) violence}]\\
        Answer index: 3 \\
    \end{minipage}}
&
    \scalebox{0.8}{\begin{minipage}{400pt}
        Subject: Social Studies \\
        Skill: southern-colonies-economy-and-slavery \\
        Description: Complete the sentence. \\The largest farms in the Southern Colonies grew cash crops. Cash crops are plants meant \_\_\_. \\
        Choices: [\textcolor{red}{(a) to be sold}, (b) to be given to the government, (c) to feed wildlife, (d) to be eaten by farmers]\\
        Answer index: 0 \\
    \end{minipage}}\\
\hline
    \end{tabular}
}
    \caption{Question examples for each skill (part 4).}
    \label{tab:example_by_skill_4}
\end{table*}
\begin{table*}
    \centering
\scalebox{0.6}{
    \begin{tabular}{|c|c|}
    \hline
    \scalebox{0.8}{\begin{minipage}{400pt}
        Subject: Social Studies \\
        Skill: ancient-china \\
        Description: For people in ancient Chinese communities, flooding around the Yellow and Yangtze Rivers was a common problem. \\But floods had both positive and negative effects for people in ancient China. Which of the following was a positive effect of flooding? \\
        Choices: [(a) Floods killed many animals who lived near the river., \textcolor{red}{(b) After a river flooded, fertile soil was left on the land.}, (c) Many crops were destroyed by the floods., (d) Floods destroyed homes located near the river.]\\
        Answer index: 1 \\
    \end{minipage}}
&
    \scalebox{0.8}{\begin{minipage}{400pt}
        Subject: Social Studies \\
        Skill: origins-of-islam \\
        Description: Muslims believe there is only one god. They call this god Allah, which means "God" in Arabic. The following passage describes a Muslim belief about how Allah revealed, or communicated, messages to his followers. Read the passage. Then answer the question below. \\We believe in Allah and [the message that] was revealed to us and what was revealed to Abraham . . . and the teachings which Allah gave to Moses and Jesus and to other prophets. \\... \\Al Imran 3:84 in Zafar Ishaq Ansari, translator, Towards Understanding the Qur'an, Volume 1. Copyright 1988 by The Islamic Foundation. \\Based on this passage, which statement do most Muslims believe is true? \\
        Choices: [(a) Allah is more powerful than the god Jews and Christians worship., \textcolor{red}{(b) Allah is the same god that Jews and Christians worship.}, (c) Allah sent messages to Abraham, but not to other prophets.]\\
        Answer index: 1 \\
    \end{minipage}}\\
\hline
    \end{tabular}
}
    \caption{Question examples for each skill (part 5).}
    \label{tab:example_by_skill_5}
\end{table*}
\begin{table*}
    \centering
\scalebox{0.6}{
    \begin{tabular}{|c|c|}
    \hline
    \scalebox{0.8}{\begin{minipage}{400pt}
        Subject: Social Studies \\
        Skill: mesopotamian-empires \\
        Description: Read some of King Hammurabi's laws. Then complete the sentence below. \\If a man put out the eye of another man, his eye shall be put out. \\... \\mina:a unit of weight for measuring money, about 1 pound \\Complete the sentence. \\According to Hammurabi's law code, if a man commits a crime against an enslaved person, his punishment would be \_\_\_ the punishment for a crime committed against his equal. \\
        Choices: [\textcolor{red}{(a) less harsh than}, (b) equally as harsh as, (c) more harsh than]\\
        Answer index: 0 \\
    \end{minipage}}
&
    \scalebox{0.8}{\begin{minipage}{400pt}
        Subject: Social Studies \\
        Skill: the-kingdoms-of-kush \\
        Description: Around 2300 BCE, the first kingdom of Kush began to develop, called the Kerman Kingdom. It grew out of a city called Kerma. At that time, Kush did not have its own written language. So today, researchers study objects and buildings instead of written sources to learn about the Kerman Kingdom. Read the description of the largest monument in the Kerman Kingdom. Then answer the question below. \\The Western Deffufa (duh-FOO-fuh) is one of the oldest monuments in Africa. It was built by the rulers of the Kerman Kingdom around 2000 BCE. When the Western Deffufa was built, it was one of the largest buildings in the region. Today, even though some of the building has crumbled away, it is still around 60 feet tall. That's about as tall as a six-story building!  \\... \\The Western Deffufa was made out of a common building material called mudbrick. Each brick had to be individually shaped and then baked in the sun. Then people needed to transport the bricks to the construction site and stack them up to make the walls of the building. This process took a lot of time. It also cost the Kerman rulers a lot of money to feed all of the workers. \\What does the Western Deffufa tell researchers about the Kerman Kingdom? \\
        Choices: [\textcolor{red}{(a) The Kerman Kingdom must have been wealthy and powerful in order to build it.}, (b) The Kerman Kingdom used a rare type of building material., (c) The Kerman Kingdom used the structure for military activities.]\\
        Answer index: 0 \\
    \end{minipage}}\\
\hline
    \end{tabular}
}
    \caption{Question examples for each skill (part 6).}
    \label{tab:example_by_skill_6}
\end{table*}
\begin{table*}
    \centering
\scalebox{0.6}{
    \begin{tabular}{|c|c|}
    \hline
    \scalebox{0.8}{\begin{minipage}{400pt}
        Subject: Social Studies \\
        Skill: benjamin-franklin \\
        Description: Benjamin Franklin played an important part in which conflict? \\
        Choices: [(a) World War I, (b) the Civil War, \textcolor{red}{(c) the American Revolution}, (d) the War of 1812]\\
        Answer index: 2 \\
    \end{minipage}}
&
    \scalebox{0.8}{\begin{minipage}{400pt}
        Subject: Social Studies \\
        Skill: ramadan \\
        Description: Complete the sentence. \\During Ramadan, Muslims remember an important event in the life of \_\_\_. \\
        Choices: [\textcolor{red}{(a) Muhammad}, (b) Pope John Paul II, (c) Guru Nanak, (d) Martin Luther]\\
        Answer index: 0 \\
    \end{minipage}}\\
\hline
    \scalebox{0.8}{\begin{minipage}{400pt}
        Subject: Social Studies \\
        Skill: dia-de-los-muertos \\
        Description: When is Día de los Muertos celebrated? \\
        Choices: [(a) April 15, (b) October 27, \textcolor{red}{(c) November 1 and 2}, (d) the last Monday in May]\\
        Answer index: 2 \\
    \end{minipage}}
&
    \scalebox{0.8}{\begin{minipage}{400pt}
        Subject: Social Studies \\
        Skill: jackie-robinson \\
        Description: Why do we look up to Jackie Robinson? \\
        Choices: [(a) He was bossy., \textcolor{red}{(b) He was brave.}, (c) He was popular., (d) He was funny.]\\
        Answer index: 1 \\
    \end{minipage}}\\
\hline

    \end{tabular}
}
    \caption{Question examples for each skill (part 7).}
    \label{tab:example_by_skill_7}
\end{table*}
\begin{table*}
    \centering
\scalebox{0.6}{
    \begin{tabular}{|c|c|}
    \hline
    \scalebox{0.8}{\begin{minipage}{400pt}
        Subject: Social Studies \\
        Skill: new-england-colonies-economy-and-interactions-with-native-americans \\
        Description: Colonists traded with Native American groups. But colonists also fought Native American groups for land and resources. For example, in the Pequot War (1636–1638), New England colonists fought the Pequot people for control of the valuable farmland and trading routes along the Connecticut River. \\The adapted passage below was written by a New England colonist who fought in the Pequot War. In this passage, the colonist describes how an army of New England colonists and Native American allies attacked the most important Pequot town. \\Many Pequot were burnt when we set fire to the town, men, women, and children, others forced out which our soldiers killed; down fell men, women, and children. Those that escaped us, fell into the hands of [the Native Americans fighting on the side of the colonists], that were behind us. There were about four hundred Pequot in this town, and not above five of them escaped out of our hands. \\Captain John Underhill, Newes from America, 1638. \\This passage describes a key event of the Pequot War. According to the passage, what happened during this event? \\
        Choices: [(a) The Pequot defeated the army of colonists and their allies., (b) The Pequot and the colonists agreed to share the Connecticut River., \textcolor{red}{(c) The colonists and their allies killed or captured most of the Pequot.}, (d) The colonists captured the Pequot warriors but did not harm the other Pequot.]\\
        Answer index: 2 \\
    \end{minipage}}
&
    \scalebox{0.8}{\begin{minipage}{400pt}
        Subject: Social Studies \\
        Skill: state-and-local-government \\
        Description: Sometimes it is not clear what powers a state government has. The passage below describes a conflict between a state government and the federal government over schools that were segregated, or divided, by race. Read the passage. Then answer the question below. \\Before 1954, many states segregated public schools based on race. In these states, African American students and white students were not allowed to go to the same schools. Opponents and supporters of segregation disagreed about whether the state government or the national government had the power to decide whether schools should be segregated: \\... \\In the case Brown v. Board of Education (1954), the U.S. Supreme Court ruled that segregated schools violated the rights of American citizens under federal law. The court ordered states to stop segregating schools. When the state government of Arkansas refused to obey the Supreme Court, the president of the United States sent soldiers to enforce the court's decision. \\Based on the passage, which part of government decides whether the Constitution allows a state government to do something? \\
        Choices: [\textcolor{red}{(a) the Supreme Court}, (b) the president, (c) state governments, (d) Congress]\\
        Answer index: 0 \\
    \end{minipage}}\\
\hline
    \end{tabular}
}
    \caption{Question examples for each skill (part 8).}
    \label{tab:example_by_skill_8}
\end{table*}
\begin{table*}
    \centering
\scalebox{0.6}{
    \begin{tabular}{|c|c|}
    \hline
    \scalebox{0.8}{\begin{minipage}{400pt}
        Subject: Social Studies \\
        Skill: the-executive-branch \\
        Description: Read Section 1 of Article II of the United States Constitution. Then answer the question. \\The executive power shall be vested in a president of the United States of America. He shall hold his office during the term of four years, and, together with the vice president, chosen for the same term, be elected. \\... \\vested in:given to \\Complete the sentence. \\According to the Constitution, the president has the power to \_\_\_. \\
        Choices: [\textcolor{red}{(a) carry out laws}, (b) decide how long a term of office is, (c) renovate the White House, (d) run elections]\\
        Answer index: 0 \\
    \end{minipage}}
&
    \scalebox{0.8}{\begin{minipage}{400pt}
        Subject: Social Studies \\
        Skill: the-revolutionary-war-conclusion-and-aftermath \\
        Description: As the war continued, many Patriots lashed out at Loyalists, or those who remained loyal to Great Britain during the Revolutionary War. Read the passage from an editorial appearing in the Pennsylvania Packet, one of the first daily newspapers in America. Then answer the question below. \\Send [the Loyalists] where they may enjoy their beloved slavery to perfection—send them to the island of Britain; there let them drink the cup of slavery and eat the bread of bitterness all the days of their existence. . .. Never let them return to this happy land—never let them taste the sweets of that independence which they strove to prevent. Banishment, perpetual banishment, should be their lot. \\What is the main argument in the passage? \\
        Choices: [(a) Patriots should take away Loyalists' right to vote., \textcolor{red}{(b) Loyalists should be banished to Britain.}, (c) Loyalists should be enslaved., (d) Patriots should convince Loyalists of the value of independence.]\\
        Answer index: 1 \\
    \end{minipage}}\\
\hline
    \end{tabular}
}
    \caption{Question examples for each skill (part 9).}
    \label{tab:example_by_skill_9}
\end{table*}
\begin{table*}
    \centering
\scalebox{0.6}{
    \begin{tabular}{|c|c|}
    \hline
    \scalebox{0.8}{\begin{minipage}{400pt}
        Subject: Social Studies \\
        Skill: causes-of-the-american-revolution-introduction-to-mercantilism-and-the-navigation-acts \\
        Description: To increase Great Britain's wealth, Parliament passed laws that gave each part of the British Empire a different economic role: \\Colonists would send raw materials to people in Great Britain. Raw materials, such as raw iron and wool, are directly from nature. \\People in Great Britain would turn the raw materials into manufactured goods, or finished products, such as fabric and tools. British merchants could sell these manufactured goods back to the colonies or to other countries. \\Complete the text. \\Parlaiment wanted the country to earn more money from selling exports than it spent buying imports. The raw materials that Great Britain imported from the colonies were \_\_\_ the manufactured goods that Great Britain exported for colonists to purchase. Parliament's policies were part of what created the system of Triangular Trade.  \\
        Choices: [(a) more valuable than, (b) as valuable as, \textcolor{red}{(c) less valuable than}]\\
        Answer index: 2 \\
    \end{minipage}}
&
    \scalebox{0.8}{\begin{minipage}{400pt}
        Subject: Social Studies \\
        Skill: the-revolutionary-war-preparing-for-war \\
        Description: After 1763, the British Parliament began to tax the Thirteen Colonies more heavily to help pay off debt from the French and Indian War. Parliament also restricted colonial trade. Many colonists resented these new taxes and restrictions. \\After the passage of the Stamp Act in 1765, which required Americans to pay a tax on all paper they used, an American colonist drew a cartoon with this caption: \\The moral is, that the colonies may be ruined, but that Britain would thereby be maimed. \\maimed:permanently wounded \\What is the author most likely arguing in the caption? \\
        Choices: [(a) The colonies are not a part of the British Empire., \textcolor{red}{(b) Great Britain is harming itself by taxing the colonies.}, (c) Great Britain's taxation policies are good for the colonies., (d) The colonies will be better off if they declare independence from Great Britain.]\\
        Answer index: 1 \\
    \end{minipage}}\\
\hline
    \end{tabular}
}
    \caption{Question examples for each skill (part 10).}
    \label{tab:example_by_skill_10}
\end{table*}
\begin{table*}
    \centering
\scalebox{0.6}{
    \begin{tabular}{|c|c|}
    \hline
    \scalebox{0.8}{\begin{minipage}{400pt}
        Subject: Social Studies \\
        Skill: industrialization-in-the-gilded-age \\
        Description: Industrialization brought dramatic changes to Americans' daily lives. Read a modern scholar's description of those changes. Then answer the question below. \\Within a few decades, urban American homes became networked. . . . Instead of relying on candles and [fuel] carried into the home, each home was connected to the electricity network that provided electric light . . . Instead of relying on . . . outhouses and cesspools, each home was gradually connected to two more networks, one bringing in a supply of clean running water and the other taking waste out into sewers. Houses of the rich after 1880 and of the working class after 1910 were increasingly supplied with central heating. \\... \\Robert J. Gordon, The Rise and Fall of American Growth, Princeton University Press, 2016. \\What is the main idea of the passage? \\
        Choices: [(a) As the U.S. industrialized, fewer homes had electricity and central heating., \textcolor{red}{(b) Industrialization gave many Americans much more comfortable lives.}, (c) Electric lighting and sewers made homes dirtier and more dangerous., (d) Factories became very modern, but regular people's houses did not change.]\\
        Answer index: 1 \\
    \end{minipage}}
&
    \scalebox{0.8}{\begin{minipage}{400pt}
        Subject: Social Studies \\
        Skill: causes-of-the-american-revolution-politics-and-society \\
        Description: Andrew Burnaby, an English minister, traveled around the Thirteen Colonies during the French and Indian War. Based on his observations, Burnaby argued that America would never be independent of Great Britain. Read Burnaby's argument. Then answer the question below. \\The difference of character, of manners, of religion, of interest, of the different colonies [is so great that if they were] left to themselves there would soon be a civil war. \\Why did Burnaby think that America would never be a nation independent from Great Britain? \\
        Choices: [(a) The colonies were united in their support of Great Britain., (b) Colonial businesses needed British customers., \textcolor{red}{(c) There were too many differences between the colonies.}, (d) The colonies signed a pledge to Great Britain after the French and Indian War.]\\
        Answer index: 2 \\
    \end{minipage}}\\
\hline
    \end{tabular}
}
    \caption{Question examples for each skill (part 11).}
    \label{tab:example_by_skill_11}
\end{table*}
\begin{table*}
    \centering
\scalebox{0.55}{
    \begin{tabular}{|c|c|}
    \hline
    \scalebox{0.8}{\begin{minipage}{400pt}
        Subject: Social Studies \\
        Skill: origins-of-hinduism \\
        Description: Many historians argue that Hinduism started to develop when Indus Valley and Aryan cultures mixed. Look at the following information about some common Hindu beliefs and religious practices, or actions, that developed over time. Then use the information to answer the question below. \\Some Hindus believe that there are hundreds of different gods, but some think there is just one god who has many different characteristics. \\... \\Hindus in different areas often celebrate different holidays and festivals. \\Based on the information above, which of the following statements is true? \\
        Choices: [\textcolor{red}{(a) There are many different ways of practicing Hinduism.}, (b) There is only one main way of practicing Hinduism.]\\
        Answer index: 0 \\
    \end{minipage}}
&
    \scalebox{0.8}{\begin{minipage}{400pt}
        Subject: Social Studies \\
        Skill: origins-of-buddhism \\
        Description: Historians agree that Buddhism began with the teachings of Siddhartha Gautama (si-DAR-tuh GOW-tuh-muh). During his life, Siddhartha became a great teacher and was given the title "buddha" (BOO-duh). \\Read about some common Buddhist beliefs. Then answer the question below. \\There were buddhas who lived before Siddhartha, and there will be buddhas who come after him. \\... \\Siddhartha is often called "the Buddha," instead of just "buddha," because he is the buddha of this age. \\What does it mean to say that Siddhartha is "the buddha of this age"? \\
        Choices: [\textcolor{red}{(a) He was the buddha who lived most recently.}, (b) He was not very old when he became a buddha., (c) He was less important than future buddhas., (d) He was the first buddha who ever lived.]\\
        Answer index: 0 \\
    \end{minipage}}\\
\hline

    \end{tabular}
}
    \caption{Question examples for each skill (part 12).}
    \label{tab:example_by_skill_12}
\end{table*}
\begin{table*}
    \centering
\scalebox{0.6}{
    \begin{tabular}{|c|c|}
    \hline
    \scalebox{0.8}{\begin{minipage}{400pt}
        Subject: Social Studies \\
        Skill: foundations-of-maya-civilization \\
        Description: Complete the text. \\Archaeologists believe that the ancestors of the Maya first arrived in Central America more than 12,000 years ago. At first, these people were \_\_\_ and traveled long distances to find food. But over time, they learned to grow crops instead. By at least 1500 BCE, they began to settle in permanent \_\_\_. This was the beginning of the Maya civilization. \\
        Choices: [(a) explorers . . . cities, (b) farmers . . . palaces, \textcolor{red}{(c) hunter-gatherers . . . villages}]\\
        Answer index: 2 \\
    \end{minipage}}
&
    \scalebox{0.8}{\begin{minipage}{400pt}
        Subject: Social Studies \\
        Skill: classical-athens-geography-and-society \\
        Description: Complete the text. \\For most of its history, ancient Athens ruled itself and was not part of a larger country. In other words, Athens was a \_\_\_. \\
        Choices: [(a) region, \textcolor{red}{(b) city-state}, (c) kingdom]\\
        Answer index: 1 \\
    \end{minipage}}\\
\hline
    \end{tabular}
}
    \caption{Question examples for each skill (part 13).}
    \label{tab:example_by_skill_13}
\end{table*}
\begin{table*}
    \centering
\scalebox{0.6}{
    \begin{tabular}{|c|c|}
    \hline
    \scalebox{0.8}{\begin{minipage}{400pt}
        Subject: Language Arts \\
        Skill: is-it-a-complete-sentence-or-a-fragment \\
        Description: Is this a complete sentence? \\Monica speaks four languages. \\
        Choices: [\textcolor{red}{(a) Monica speaks four languages.}]\\
        Answer index: 0 \\
    \end{minipage}}
&
    \scalebox{0.8}{\begin{minipage}{400pt}
        Subject: Language Arts \\
        Skill: is-it-a-phrase-or-a-clause \\
        Description: Is the group of words in bold a phrase or a clause? \\**Emily's suitcase was too large to fit in the overhead bin**, so unfortunately she had to check it. \\
        Choices: [(a) phrase, \textcolor{red}{(b) clause}]\\
        Answer index: 1 \\
    \end{minipage}}\\
\hline
    \scalebox{0.8}{\begin{minipage}{400pt}
        Subject: Language Arts \\
        Skill: choose-between-subject-and-object-pronouns \\
        Description: Select the correct pronoun to complete the sentence. \\"Thanks to your efforts, \_\_\_ have emerged as leaders in the energy industry," Mr. Bullock announced at the company meeting. \\
        Choices: [(a) us, \textcolor{red}{(b) we}]\\
        Answer index: 1 \\
    \end{minipage}}
&
    \scalebox{0.8}{\begin{minipage}{400pt}
        Subject: Language Arts \\
        Skill: is-the-sentence-simple-compound-complex-or-compound-complex \\
        Description: Which type of sentence is this? \\Henry always approaches difficult tasks enthusiastically, and he frequently motivates others with his energy and fervor. \\
        Choices: [(a) simple, \textcolor{red}{(b) compound}, (c) complex, (d) compound-complex]\\
        Answer index: 1 \\
    \end{minipage}}\\
\hline
    \end{tabular}
}
    \caption{Question examples for each skill (part 14).}
    \label{tab:example_by_skill_14}
\end{table*}
\begin{table*}
    \centering
\scalebox{0.6}{
    \begin{tabular}{|c|c|}
    \hline
    \scalebox{0.8}{\begin{minipage}{400pt}
        Subject: Language Arts \\
        Skill: revise-the-sentence-using-a-stronger-verb \\
        Description: Read the sentence. \\After a difficult climb, Madelyn **went** back down the hill. \\Now, revise the sentence to show that Madelyn was **clumsy**. \\After a difficult climb, Madelyn \_\_\_ back down the hill. \\
        Choices: [(a) floated, \textcolor{red}{(b) stumbled}, (c) plunged]\\
        Answer index: 1 \\
    \end{minipage}}
&
    \scalebox{0.8}{\begin{minipage}{400pt}
        Subject: Language Arts \\
        Skill: use-the-correct-subject-or-verb \\
        Description: Complete the sentence with the best **verb**. \\Both of us \_\_\_ the banjo, but Jenna also plays the flute. \\
        Choices: [(a) plays, \textcolor{red}{(b) play}]\\
        Answer index: 1 \\
    \end{minipage}}\\
\hline
    \scalebox{0.8}{\begin{minipage}{400pt}
        Subject: Language Arts \\
        Skill: choose-between-personal-and-reflexive-pronouns \\
        Description: Select the best pronoun to complete the sentence. \\Will you please give \_\_\_ directions to the Lanberry Community Center? \\
        Choices: [\textcolor{red}{(a) me}, (b) myself]\\
        Answer index: 0 \\
    \end{minipage}}
&
    \scalebox{0.8}{\begin{minipage}{400pt}
        Subject: Language Arts \\
        Skill: choose-between-subject-and-object-personal-pronouns \\
        Description: Select the correct pronoun to complete the sentence. \\When Tara is upset, she always confides in \_\_\_.  \\
        Choices: [(a) he, \textcolor{red}{(b) him}]\\
        Answer index: 1 \\
    \end{minipage}}\\
\hline
    \end{tabular}
}
    \caption{Question examples for each skill (part 15).}
    \label{tab:example_by_skill_15}
\end{table*}
\begin{table*}
    \centering
\scalebox{0.6}{
    \begin{tabular}{|c|c|}
    \hline
    \scalebox{0.8}{\begin{minipage}{400pt}
        Subject: Language Arts \\
        Skill: identify-text-structures \\
        Description: Read the text. \\Although there are many kinds of writing, most people divide texts into two different types: fiction and nonfiction. Fiction means stories that are made up. "Charlotte's Web" is fiction because, of course, spiders and pigs don't really talk. Nonfiction, however, gives you facts about the real world. You can find nonfiction books about history, science, real people, animals, and many other topics. Nonfiction also includes newspapers and textbooks, while fiction includes stories, plays, and novels. Reading both fiction and nonfiction can be fun and educational. \\Which text structure does the text use? \\
        Choices: [(a) cause-effect, \textcolor{red}{(b) compare-contrast}]\\
        Answer index: 1 \\
    \end{minipage}}
&
    \scalebox{0.8}{\begin{minipage}{400pt}
        Subject: Language Arts \\
        Skill: which-sentence-matches-the-definition \\
        Description: Which sentence uses this meaning of **concrete**? \\**concrete** "adjective" specific and real \\
        Choices: [\textcolor{red}{(a) His fingerprints were **concrete** proof that he had opened the safe.}, (b) We can't walk on the sidewalk because the **concrete** is still drying.]\\
        Answer index: 0 \\
    \end{minipage}}\\
\hline

    \end{tabular}
}
    \caption{Question examples for each skill (part 16).}
    \label{tab:example_by_skill_16}
\end{table*}
\begin{table*}
    \centering
\scalebox{0.6}{
    \begin{tabular}{|c|c|}
    \hline
    \scalebox{0.8}{\begin{minipage}{400pt}
        Subject: Language Arts \\
        Skill: draw-inferences-from-a-text \\
        Description: Read the text. \\Kari looked under the table. Then she searched the whole room. \\"You have got to be kidding!" Kari shouted. \\Grumbling, Kari tore the puzzle apart and put it back in the box. On the top of the box, she crossed out 1000 and wrote 999 above it. \\Based on the text, which sentence is more likely to be true? \\
        Choices: [(a) Kari doesn't like building puzzles., \textcolor{red}{(b) Kari's puzzle is missing a piece.}]\\
        Answer index: 1 \\
    \end{minipage}}
&
    \scalebox{0.8}{\begin{minipage}{400pt}
        Subject: Language Arts \\
        Skill: is-the-sentence-simple-or-compound \\
        Description: Which is a **compound sentence**? \\
        Choices: [\textcolor{red}{(a) Jayden grew too many pumpkins, so he sold the extras at the market.}, (b) British general John Burgoyne surrendered to American forces at the Battle of Saratoga.]\\
        Answer index: 0 \\
    \end{minipage}}\\
\hline
    \end{tabular}
}
    \caption{Question examples for each skill (part 17).}
    \label{tab:example_by_skill_17}
\end{table*}
\begin{table*}
    \centering
\scalebox{0.6}{
    \begin{tabular}{|c|c|}
    \hline
    \scalebox{0.8}{\begin{minipage}{400pt}
        Subject: Language Arts \\
        Skill: which-word-is-not-like-the-others \\
        Description: Which word is not like the others? \\
        Choices: [(a) big, (b) tall, (c) little, \textcolor{red}{(d) nice}]\\
        Answer index: 3 \\
    \end{minipage}}
&
    \scalebox{0.8}{\begin{minipage}{400pt}
        Subject: Language Arts \\
        Skill: complete-the-sentence-with-the-correct-sight-word-sets-1-2-3 \\
        Description: Complete the sentence. \\I did not want to trick \_\_\_. \\
        Choices: [(a) has, (b) as, \textcolor{red}{(c) them}]\\
        Answer index: 2 \\
    \end{minipage}}\\
\hline
    \scalebox{0.8}{\begin{minipage}{400pt}
        Subject: Language Arts \\
        Skill: complete-the-sentence-with-the-correct-sight-word-sets-4-5-6-7 \\
        Description: Complete the sentence. \\I ate a hot dog \_\_\_ went to bed. \\
        Choices: [(a) four, (b) more, \textcolor{red}{(c) then}]\\
        Answer index: 2 \\
    \end{minipage}}
&
    \scalebox{0.8}{\begin{minipage}{400pt}
        Subject: Language Arts \\
        Skill: use-the-correct-modal-verb \\
        Description: Select the modal verb that best completes the sentence. \\The coach \_\_\_ let Mark join the team if it weren't so late in the season. \\
        Choices: [\textcolor{red}{(a) would}, (b) will]\\
        Answer index: 0 \\
    \end{minipage}}\\
\hline
    \end{tabular}
}
    \caption{Question examples for each skill (part 18).}
    \label{tab:example_by_skill_18}
\end{table*}
\begin{table*}
    \centering
\scalebox{0.6}{
    \begin{tabular}{|c|c|}
    \hline
    \scalebox{0.8}{\begin{minipage}{400pt}
        Subject: Language Arts \\
        Skill: words-with-able-and-ible \\
        Description: Which item is not a **collectible**? \\
        Choices: [\textcolor{red}{(a) an ordinary spiral notebook}, (b) a rare comic book]\\
        Answer index: 0 \\
    \end{minipage}}
&
    \scalebox{0.8}{\begin{minipage}{400pt}
        Subject: Language Arts \\
        Skill: use-the-correct-frequently-confused-word \\
        Description: Complete the text with the correct word. \\Rhianna made a wire sculpture of a tree for her art project. She used steel wool to make \_\_\_ leaves. \\
        Choices: [\textcolor{red}{(a) its}, (b) it's]\\
        Answer index: 0 \\
    \end{minipage}}\\
\hline
    \scalebox{0.8}{\begin{minipage}{400pt}
        Subject: Language Arts \\
        Skill: words-with-mis \\
        Description: Which type of behavior shouldn't cause **mistrust**? \\
        Choices: [(a) suspicious behavior, \textcolor{red}{(b) honest behavior}]\\
        Answer index: 1 \\
    \end{minipage}}
&
    \scalebox{0.8}{\begin{minipage}{400pt}
        Subject: Language Arts \\
        Skill: which-sentence-is-more-formal \\
        Description: Which sentence is more formal? \\
        Choices: [\textcolor{red}{(a) Visitors are advised that admission to the aircraft museum has been raised by two dollars.}, (b) FYI, admission to the aircraft museum has been raised by two dollars.]\\
        Answer index: 0 \\
    \end{minipage}}\\
\hline
    \end{tabular}
}
    \caption{Question examples for each skill (part 19).}
    \label{tab:example_by_skill_19}
\end{table*}
\begin{table*}
    \centering
\scalebox{0.6}{
    \begin{tabular}{|c|c|}
    \hline
    \scalebox{0.8}{\begin{minipage}{400pt}
        Subject: Language Arts \\
        Skill: identify-the-irregular-past-tense-ii \\
        Description: Which sentence is in the past tense? \\
        Choices: [(a) Susan **feeds** her bird., \textcolor{red}{(b) Susan **fed** her bird.}]\\
        Answer index: 1 \\
    \end{minipage}}
&
    \scalebox{0.8}{\begin{minipage}{400pt}
        Subject: Language Arts \\
        Skill: which-sentence-is-in-the-regular-past-tense \\
        Description: Which sentence is in the past tense? \\
        Choices: [\textcolor{red}{(a) Cassie **acted** happy.}, (b) We **discover** hidden treasures.]\\
        Answer index: 0 \\
    \end{minipage}}\\
\hline
    \scalebox{0.8}{\begin{minipage}{400pt}
        Subject: Language Arts \\
        Skill: which-sentence-uses-an-antonym \\
        Description: Read the sentence. Select the sentence that uses an antonym in place of the word in bold. \\Bella's dad has always been **short**. \\
        Choices: [\textcolor{red}{(a) Bella's dad has always been **tall**.}, (b) Bella's dad has always been **shy**.]\\
        Answer index: 0 \\
    \end{minipage}}
&
    \scalebox{0.8}{\begin{minipage}{400pt}
        Subject: Language Arts \\
        Skill: use-action-verbs \\
        Description: Complete the sentence with the action verb. \\Mrs. Carlson \_\_\_ nuts into the cookie batter. \\
        Choices: [\textcolor{red}{(a) mixes}, (b) butter]\\
        Answer index: 0 \\
    \end{minipage}}\\
\hline
    \end{tabular}
}
    \caption{Question examples for each skill (part 20).}
    \label{tab:example_by_skill_20}
\end{table*}
\begin{table*}
    \centering
\scalebox{0.6}{
    \begin{tabular}{|c|c|}
    \hline
    \scalebox{0.8}{\begin{minipage}{400pt}
        Subject: Language Arts \\
        Skill: read-and-understand-informational-passages \\
        Description: Review the **seventh paragraph**. \\title: Meditating in School \\... \\This is where the need for good, sound research comes in. Past studies showing that mindfulness helps medical patients are not enough. To really determine whether mindfulness helps improve grades, research needs to focus specifically on mindfulness in schools—and what impact, if any, it has. That way, educators can make informed decisions about where to put their resources. Until then, whether mindfulness is included in classroom programming will remain the choice of individual teachers. \\Based on the text, what still needs to be determined about mindfulness in schools? \\
        Choices: [(a) whether meditation affects different students in different ways, (b) how to raise the funds needed to teach meditation in schools, \textcolor{red}{(c) whether mindfulness meditation has a positive impact in a school setting}]\\
        Answer index: 2 \\
    \end{minipage}}
&
    \scalebox{0.8}{\begin{minipage}{400pt}
        Subject: Language Arts \\
        Skill: choose-evidence-to-support-a-claim \\
        Description: Read the claim below. \\Gardeners should never use pesticides on their plants. \\Select the piece of evidence that best supports this claim. \\
        Choices: [(a) Organic pesticides tend to cost more than conventional ones, and they may be less effective., \textcolor{red}{(b) Pets and children often play in and around gardens, where they can be exposed to pesticides.}]\\
        Answer index: 1 \\
    \end{minipage}}\\
\hline
    \end{tabular}
}
    \caption{Question examples for each skill (part 21).}
    \label{tab:example_by_skill_21}
\end{table*}
\begin{table*}
    \centering
\scalebox{0.6}{
    \begin{tabular}{|c|c|}
    \hline
    \scalebox{0.8}{\begin{minipage}{400pt}
        Subject: Language Arts \\
        Skill: identify-the-authors-purpose \\
        Description: Review the text. \\Good study habits train your brain for success. Applying effective learning methods will not only improve your grades but, more importantly, increase your long-term knowledge. \\Cramming at the last minute is ineffective. Instead, try to create a study routine. Schedule specific blocks of time to study every day. Next, find the best environment that helps you concentrate. Some students do better with the silence of a library, while others need the background noise or music of a public space. \\Now that you have a time and a place, you need a strategy. Look at your tasks and make a list. Setting small, manageable goals for your study session will help you work efficiently. Do the most difficult assignments first, while your mind is fresh. Finally, do not be afraid to take limited breaks to restore your motivation. \\Which best shows that the author's purpose is "to explain some productive study techniques to readers"? \\
        Choices: [(a) The author argues that students need to work in a place with "background noise" in order to study efficiently., \textcolor{red}{(b) The author outlines a series of best practices to follow to improve study habits.}]\\
        Answer index: 1 \\
    \end{minipage}}
&
    \scalebox{0.8}{\begin{minipage}{400pt}
        Subject: Language Arts \\
        Skill: is-it-a-direct-object-or-an-indirect-object \\
        Description: Is the phrase in bold a direct object or an indirect object? \\A guard denied **the tourists** entrance to the palace. \\
        Choices: [(a) direct object, \textcolor{red}{(b) indirect object}]\\
        Answer index: 1 \\
    \end{minipage}}\\
\hline
    \end{tabular}
}
    \caption{Question examples for each skill (part 22).}
    \label{tab:example_by_skill_22}
\end{table*}
\begin{table*}
    \centering
\scalebox{0.6}{
    \begin{tabular}{|c|c|}
    \hline
    \scalebox{0.8}{\begin{minipage}{400pt}
        Subject: Language Arts \\
        Skill: analyze-passages-from-narrative-of-the-life-of-frederick-douglass-part-1 \\
        Description: Read the following passage from "Narrative of the Life of Frederick Douglass". In this excerpt, Douglass describes some features found on the plantation where he was enslaved as a young child. \\Colonel Lloyd kept a large and finely cultivated garden, which afforded almost constant employment for four men, besides the chief gardener, Mr. M'Durmond. This garden was probably the greatest attraction of the place. During the summer months, people came from far and near—from Baltimore, Easton, and Annapolis—to see it. It abounded in fruits of almost every description, from the hardy apple of the north to the delicate orange of the south. This garden was not the least source of trouble on the plantation. Its excellent fruit was quite a temptation to the hungry swarms of boys, as well as the older slaves, belonging to the colonel, few of whom had the virtue or the vice to resist it. \\... \\From Frederick Douglass, Narrative of the Life of Frederick Douglass \\What is this passage mainly about? \\
        Choices: [(a) It is about why Douglass enjoyed spending time in the garden and stable on Colonel Lloyd's plantation., \textcolor{red}{(b) It is about how Colonel Lloyd's fancy garden and fine horses created challenges for enslaved people.}, (c) It is about how Colonel Lloyd paid more attention to his garden than to the horses in his stable., (d) It is about why enslaved people preferred to work in Colonel Lloyd's stable instead of his garden.]\\
        Answer index: 1 \\
    \end{minipage}}
&
    \scalebox{0.8}{\begin{minipage}{400pt}
        Subject: Language Arts \\
        Skill: analyze-short-stories \\
        Description: Read the story. \\title: Chiarelli's Bakery \\... \\Ms. Chiarelli nodded impatiently. "Come on! If we hurry, we can make cassatelle too." \\Based on the **fourth paragraph**, what is Aram's attitude about his future career? \\
        Choices: [\textcolor{red}{(a) He is confident that he will achieve success.}, (b) He is sure that he will fail, but he wants to try., (c) He is uncertain how things will turn out.]\\
        Answer index: 0 \\
    \end{minipage}}\\
\hline
    \end{tabular}
}
    \caption{Question examples for each skill (part 23).}
    \label{tab:example_by_skill_23}
\end{table*}
\begin{table*}
    \centering
\scalebox{0.6}{
    \begin{tabular}{|c|c|}
    \hline
    \scalebox{0.8}{\begin{minipage}{400pt}
        Subject: Language Arts \\
        Skill: use-context-as-a-clue-to-the-meanings-of-foreign-expressions \\
        Description: What is the meaning of the foreign expression in bold? \\In 2014, giant moths swarmed **en masse** in parts of Southeast Asia. More than eight hundred sightings were reported in Singapore, while thousands disrupted a soccer match at Darul Makmur Stadium in Malaysia. \\
        Choices: [(a) coincidentally, \textcolor{red}{(b) as a group}, (c) unexpectedly, (d) fatally]\\
        Answer index: 1 \\
    \end{minipage}}
&
    \scalebox{0.8}{\begin{minipage}{400pt}
        Subject: Language Arts \\
        Skill: use-the-correct-foreign-expression \\
        Description: Complete the text with the correct foreign expression. \\The Norwood Bistro features an eclectic \_\_\_ menu so that diners can select a variety of dishes. \\
        Choices: [\textcolor{red}{(a) à la carte}, (b) crème de la crème]\\
        Answer index: 0 \\
    \end{minipage}}\\
\hline
    \end{tabular}
}
    \caption{Question examples for each skill (part 24).}
    \label{tab:example_by_skill_24}
\end{table*}
\begin{table*}
    \centering
\scalebox{0.6}{
    \begin{tabular}{|c|c|}
    \hline
    \scalebox{0.8}{\begin{minipage}{400pt}
        Subject: Language Arts \\
        Skill: analyze-short-stories-set-2 \\
        Description: Read the story. \\title: Water Rising \\... \\Polara looked to the horizon, at the unmistakable sparkle of the approaching sea. She wasn't sure how brave she felt, but she was willing. "I am," she answered. \\Based on the **first and second paragraphs**, what internal conflict does Polara struggle with? \\
        Choices: [(a) the duty to stand watch versus her inclination to daydream, \textcolor{red}{(b) the pull of the unfamiliar versus her desire to stay at Shublik}, (c) respect for Commander Axel versus her irritation at his order, (d) the enjoyment of the view versus her fear about the water on the horizon]\\
        Answer index: 1 \\
    \end{minipage}}
&
    \scalebox{0.8}{\begin{minipage}{400pt}
        Subject: Language Arts \\
        Skill: use-appeals-to-ethos-pathos-and-logos-in-persuasive-writing \\
        Description: Which statement primarily appeals to **logos**, or reason, to support the position that **more money does not always lead to increased happiness**? \\
        Choices: [(a) Like most people, I enjoy the feeling of buying a new gadget, but it turns out that the happiness that we get from a new purchase doesn't last long., \textcolor{red}{(b) While it's true that a lack of money can cause stress and unhappiness, it does not necessarily follow that money will lead to more happiness after basic needs are met.}]\\
        Answer index: 1 \\
    \end{minipage}}\\
\hline
    \end{tabular}
}
    \caption{Question examples for each skill (part 25).}
    \label{tab:example_by_skill_25}
\end{table*}
\begin{table*}
    \centering
\scalebox{0.6}{
    \begin{tabular}{|c|c|}
    \hline
    \scalebox{0.8}{\begin{minipage}{400pt}
        Subject: Language Arts \\
        Skill: analyze-passages-from-harriet-tubman-conductor-on-the-underground-railroad-part-2 \\
        Description: Review the passage. \\In the fall of 1854, Harriet Tubman began to feel uneasy about three of her brothers. Benjamin, John and William Henry were still in Maryland, working on plantations where they had been hired out. She kept having dreams about them, vivid dreams in which she saw them sold and sent away in a chain gang. She decided to tell them that she was coming to Maryland that fall, so that they would be ready to go North with her. \\... \\Adapted from Ann Petry, Harriet Tubman: Conductor on the Underground Railroad. Copyright 1955 by Ann Petry \\What is the meaning of the word **eluded** as used in the passage? \\
        Choices: [\textcolor{red}{(a) escaped the understanding of}, (b) caused excitement about, (c) raised awareness for]\\
        Answer index: 0 \\
    \end{minipage}}
&
    \scalebox{0.8}{\begin{minipage}{400pt}
        Subject: Language Arts \\
        Skill: classify-the-figure-of-speech-euphemism-hyperbole-oxymoron-paradox \\
        Description: Which figure of speech is used in this text? \\Lauren asked her mother if she could adopt a cat, and her mother replied, "It's a **definite maybe**," so Lauren didn't want to get her hopes up. \\
        Choices: [(a) euphemism, \textcolor{red}{(b) oxymoron}]\\
        Answer index: 1 \\
    \end{minipage}}\\
\hline
    \end{tabular}
}
    \caption{Question examples for each skill (part 26).}
    \label{tab:example_by_skill_26}
\end{table*}

\end{document}